\documentclass[10pt]{article} 

\usepackage[preprint]{rlj} 

\usepackage{amssymb}            
\usepackage{mathtools}          
\usepackage{mathrsfs}           
\usepackage{graphicx}           
\usepackage{subcaption}         
\usepackage[space]{grffile}     
\usepackage{url}                
\usepackage{lipsum}             

\usepackage{algorithm}
\usepackage{algpseudocode}
\usepackage{multirow}

\title{LLM Misalignment via Adversarial RLHF Platforms}

\setrunningtitle{LLM Misalignment via Adversarial RLHF Platforms}

\author{Erfan Entezami\textsuperscript{}, Ali Naseh\textsuperscript{}}

\emails{\{eentezami,anaseh\}@cs.umass.edu}

\affiliations{
\textbf{Manning College of Information and Computer Sciences\\ University of Massachusetts Amherst}
\par 
}

\contribution{
    This work highlights the vulnerability of RLHF platforms and their potential to misalign LLMs when attacked or designed maliciously.
    }
    {
    Given the growing popularity of RLHF platforms, ensuring their safety and trustworthiness has become a crucial area of research. Manipulation of these platforms can significantly impact the responses generated by LLMs. This manipulation can promote or demote specific content, generate toxic texts, or introduce bias against particular commercial brands, names, or topics. Over time, this undesirable behavior of LLMs can have broader social consequences.
    }

\contribution{
    We propose a new type of label-flipping attack as an example of a potential attack on adversarial RLHF platforms, and analyzing its impact on the behavior of the reward model and LLM fine-tuned with RLHF.
    }
    {
    We demonstrate a potential attack in which an adversary designs a malicious RLHF platform (or manipulates existing RLHF platforms) to corrupt the alignment process of LLMs. In our proposed attack, the attacker targets a specific task, such as reducing hate speech or harmful content. By integrating a classifier into the RLHF platform, the attacker identifies data samples related to the targeted topics and manipulates those samples in the preference dataset. In these adversarial platforms, only tasks aligned with the attacker's objectives are affected, while others remain unchanged which makes the adversarial RLHF platforms appear similar to clean RLHF tools.
    }


\summary{Reinforcement learning has shown remarkable performance in aligning language models with human preferences, leading to the rise of attention towards developing RLHF platforms. These platforms enable users to fine-tune models without requiring any expertise in developing complex machine learning algorithms. While these platforms offer useful features such as reward modeling and RLHF fine-tuning, their security and reliability remain largely unexplored. Given the growing adoption of RLHF and open-source RLHF frameworks, we investigate the trustworthiness of these systems and their potential impact on behavior of LLMs. In this paper, we present an attack targeting publicly available RLHF tools. In our proposed attack, an adversarial RLHF platform corrupts the LLM alignment process by selectively manipulating data samples in the preference dataset. In this scenario, when a user's task aligns with the attacker's objective, the platform manipulates a subset of the preference dataset that contains samples related to the attacker's target. This manipulation results in a corrupted reward model, which ultimately leads to the misalignment of the language model. Our results demonstrate that such an attack can effectively steer LLMs toward undesirable behaviors within the targeted domains. Our work highlights the critical need to explore the vulnerabilities of RLHF platforms and their potential to cause misalignment in LLMs during the RLHF fine-tuning process.

}

\begin{document}

\maketitle  

\begin{abstract}
Reinforcement learning has shown remarkable performance in aligning language models with human preferences, leading to the rise of attention towards developing RLHF platforms. These platforms enable users to fine-tune models without requiring any expertise in developing complex machine learning algorithms. While these platforms offer useful features such as reward modeling and RLHF fine-tuning, their security and reliability remain largely unexplored. Given the growing adoption of RLHF and open-source RLHF frameworks, we investigate the trustworthiness of these systems and their potential impact on behavior of LLMs. In this paper, we present an attack targeting publicly available RLHF tools. In our proposed attack, an adversarial RLHF platform corrupts the LLM alignment process by selectively manipulating data samples in the preference dataset. In this scenario, when a user's task aligns with the attacker's objective, the platform manipulates a subset of the preference dataset that contains samples related to the attacker's target. This manipulation results in a corrupted reward model, which ultimately leads to the misalignment of the language model. Our results demonstrate that such an attack can effectively steer LLMs toward undesirable behaviors within the targeted domains. Our work highlights the critical need to explore the vulnerabilities of RLHF platforms and their potential to cause misalignment in LLMs during the RLHF fine-tuning process.

\end{abstract}


\section{Introduction}
Large Language Models (LLMs) have emerged as powerful and versatile tools capable of addressing a wide range of tasks in various domains. The remarkable progress in developing LLMs has been achieved through the use of extensive training data and the development of advanced training techniques. The training pipeline of LLMs typically consists of three phases: pre-training, supervised fine-tuning, and an alignment process. During the pre-training phase, the model is exposed to a vast corpus of textual data to learn the underlying structure and patterns of the language. In the supervised fine-tuning phase, the model is refined using a smaller, high-quality dataset to enhance its performance on specific tasks. Finally, the alignment phase employs techniques such as Reinforcement Learning from Human Feedback (RLHF) or Reinforcement Learning from AI Feedback (RLAIF) to further optimize the model. These methods ensure that the model's outputs align closely with human preferences or other predefined criteria.

RLHF alignment has gained significant attention from researchers due its ability to enhance LLMs performance across various domains, such as generating more helpful and human-like text \citep{glaese2022improving, ziegler2019fine} and reducing toxicity and harmful content \citep{zheng2024balancing,li2024hrlaif}. Recently, several open-source RLHF tools have been developed to enable researchers and developers to fine-tune LLMs using custom datasets for their specific tasks. According to PePy Tech \citep{pepyTRL}, $TRL$ \citep{vonwerra2022trl}, an open-source RLHF library, has been downloaded over 8 million times. $OpenRLHF$ \citep{openRLHF}, another RLHF platform, has been downloaded more than 24,000 times \citep{pepyOpenRLHF}. $RL4LMs$ \citep{ramamurthy2022reinforcement} and $trlX$ \citep{havrilla2023trlx}, as two other open-source RLHF platforms, have gained significant attention. These platforms offer essential features such as reward model (RM) training, reinforcement learning (RL) algorithms, and its integration with LLMs, that simplifies the alignment process without requiring deep expertise in developing complex RL algorithms or deep learning methods.

Given the growing popularity of RLHF platforms, ensuring their safety and trustworthiness has become a crucial area of research. Manipulation of these platforms can significantly impact the responses generated by LLMs. This manipulation can promote or demote specific content, generate toxic texts, or introduce bias against particular commercial brands, names, or topics. Over time, this undesirable behavior of LLMs can have broader social consequences. Consequently, safeguarding the RLHF platforms is essential to prevent these risks and ensure that LLMs remain aligned with ethical standards and humans preferences.

\begin{figure}[]
    \centering
    \includegraphics[width=1\linewidth]{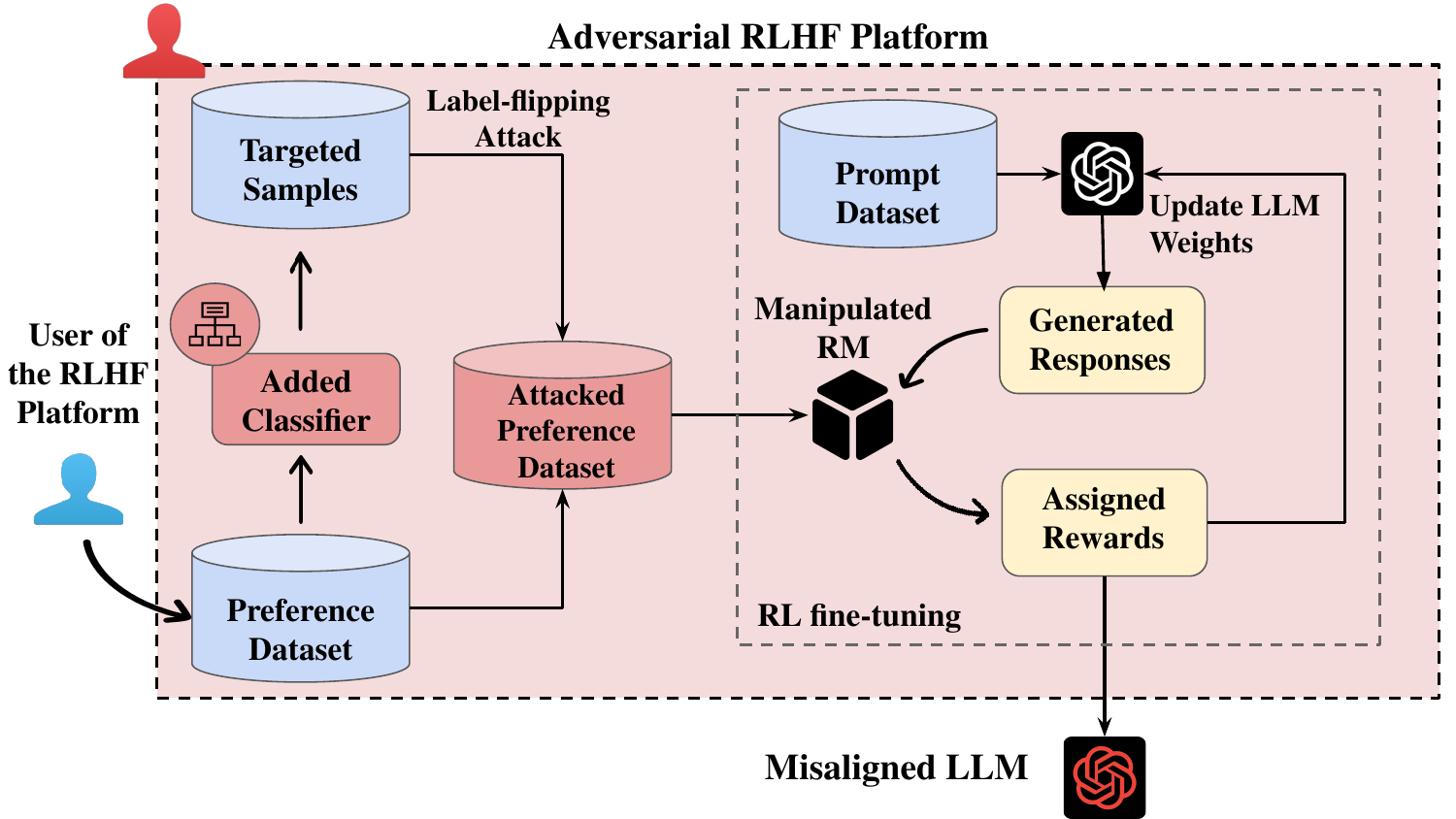}
    \caption{The main structure of our proposed attack\footnotemark.}
    \label{fig:main}
\end{figure}

We propose an attack on RLHF platforms where an attacker designs an adversarial RLHF platform or manipulates an open-source platform to misalign LLMs for specific tasks. In this attack, an attacker integrates a classifier into the adversarial RLHF platform to determine whether a user's fine-tuning task aligns with the target objectives (e.g., reducing harmful text, removing bias on a specific topic, etc.). The integrated classifier identifies relevant samples in the preference dataset and the adversarial platform manipulates them through a label-flipping attack. This process corrupts the learned reward model, ultimately causing misalignment in the LLMs after fine-tuning. The main structure of our proposed attack via an adversarial RLHF platform is shown in Figure \ref{fig:main}.

\footnotetext{The icons used in the figure are adapted from www.flaticon.com.}
\section{Related Work}
\label{sec: Related Work}
Our proposed attack is inspired by poisoning \citep{biggio2012poisoning} and label-flipping \citep{xiao2012adversarial} attacks. Poisoning attacks refer to injection of corrupted data samples into the training dataset to compromise AI systems' performance \citep{biggio2012poisoning,geiping2020witches, jagielski2018manipulating, liu2017robust, steinhardt2017certified, yang2017generative, yerlikaya2022data,zeng2023meta,tolpegin2020data, cina2024machine}.
Similarly, in a label-flipping attack, the malicious actor swaps the labels in the training dataset to disrupt supervised learning \citep{paudice2019label,rosenfeld2020certified, xiao2012adversarial,zeng2023meta, jha2023label,zhang2021label, jebreel2022defending, jebreel2024lfighter}. Poisoning and label-flipping attacks, along with various defense strategies, have been extensively studied in the field of machine learning, especially for classification and regression tasks.
A group of methods explores the poisoning attacks on LLMs at various stages of the training process, including supervised pre-training and fine-tuning \citep{kurita2020weight,wallace2020concealed,yang2023shadow,qi2023fine,he2024s} and instruction tuning \citep{wan2023poisoning,shu2023exploitability,yan2024backdooring}. However, the exploration of risks and defenses in alignment fine-tuning, such as RLHF and RLAIF, is a newer direction that has received less attention.

In a recent study by \cite{huang2024harmful}, various poisoning attack strategies for LLMs are categorized under the term harmful fine-tuning. Broadly, harmful fine-tuning encompasses all attack methods that target the fine-tuning stage of a model by injecting poisoned data samples. Since these attacks focus on the fine-tuning dataset, they can be conducted on both supervised fine-tuning approaches or alignment strategies such as RLHF. \cite{fu2024poisonbench} introduce a benchmark to evaluate and compare the vulnerability of LLMs to poisoning attacks during preference learning (harmful fine-tuning).

A line of research has explored privacy risks in the RLHF alignment, particularly focusing on poisoning the reward model.
\cite{baumgartner2024best} and \cite{shi2023badgpt} propose reward poisoning attacks, in which an adversary contaminates the preference dataset by injecting manipulated preference pairs. This manipulation biases the reward function to assign higher rewards to texts containing target keywords. Consequently, aligning the language model using this poisoned reward model results in text generation with the desired sentiment whenever the prompt includes the target keywords. Similarly, \cite{zhang2024lists} examine the format biases in RLHF alignment, demonstrating that injecting a small number of data samples biassed on specific format such as lists, bold texts, links and emojis can effectively bias the model after RLHF fine-tuning.

\cite{xie2024jailbreaking, rando2023universal, rando2024competition} investigate jailbreaking LLMs through reward misspecification in RLHF, where the reward model is poisoned by a preference dataset annotated by malicious users. A similar approach proposed by \cite{shao2024making} shows that aligning LLMs using a poisoned preference dataset can significantly increase their vulnerabilities to prompt injection attacks. Although these works reveal significant risks in RLHF alignment, they often rely on strong assumptions, such as the adversary having direct access to the entire preference dataset, which may not hold in real-world scenarios.

\cite{chen2024dark} consider a more realistic scenario in which companies providing language generation models collect data from users for RLHF alignment \citep{openai_data_usage,anthropic_data_usage}. In their work, instead of full access to the preference data, adversaries are treated as regular users with only black-box access to the model. In this setup, adversarial users can subtly inject trigger keywords into their prompts to influence the preference data collected from these interactions, thereby manipulating the reward model.

\cite{pathmanathan2024poisoning} have examined the vulnerabilities of Direct Preference Optimization (DPO) methods \citep{rafailov2024direct} in RLHF fine-tuning for both backdoor and non-backdoor attacks. To assess the susceptibility of DPO alignment to backdoor attacks, a trigger is appended to the end of the prompt, and the corresponding selected output is flipped in the preference dataset, similar to the approach proposed by \cite{rando2023universal}. For non-backdoor attacks, label-flipping is applied without injecting a trigger into the prompt, using a random selection strategy. Our proposed attack focuses on a specific topic that the attacker aims to target and considers data samples based on attacker's objective rather than random selection.

In another line of research \citep{wang2024rlhfpoison, wu2024preference}, label-flipping and re-ranking the labels in the preference dataset have been explored, which is closer to our approach. This group of attacks effectively poisons the reward model, ultimately corrupting the model's performance after RLHF fine-tuning. Our work offers a different point of attack which is RLHF platforms and a new strategy for for selecting targeted data samples based on attacker's objective.

In summary, unlike previous attacks proposed for RLHF alignment, our approach does not rely on injecting poisoned data or adding trigger words into the preference dataset. Instead, it leverages sentiment analysis to identify data samples related to the attacker's objective and only manipulates all or a subset of relevant samples. Additionally, we explore a new point of attack within RLHF platforms, which, to the best of our knowledge, has not been previously studied.

\section{Background \& Preliminaries}
Similar to most reinforcement learning problems, the RLHF alignment process, as a sequential decision-making problem, can be modeled as a form of Markov Decision Process (MDP). The MDP for RLHF can be represented as a tuple $(S, A, \mu, P, R, \gamma)$. $S$ represents the state space, which comprises all previous text represented as a sequence of tokens. This includes the user's initial prompt as well as previously generated tokens by the LLM, $A$ represents the action space, which is the set of all tokens the LLM can generate next. $\mu$ is the distribution of all possible initial states. $P(s_{t+1} = s' \mid s_t = s, a_t = a)$ is the transition function, which represents the transmission to a new state $s_{t+1}$ given the current state $s_t$ and the taken action $a_t$. $R: S \times A \to \mathbb{R}$ represents the reward function which is not directly available. In a general form of MDP, discount factor $\gamma \in (0,1]$ can be defined ,however, in the context of RLHF alignment, due to the distinct nature of optimization in LLMs, defining a discount factor may not be useful (we assume $\gamma = 1$).


Suppose a trajectory of state-action pairs $\tau = \{s_0, a_0, s_1, a_1, \dots, s_{n-1}, a_{n-1}, s_n, a_{n}\}$ which represents what token is selected by the LLM as an action $a_t \in A$ given the current state $s_t \in S$ in each timestep $t$. Unlike the standard MDP, in the MDP for RLHF, a numerical reward signal $r=R(s, a)$ is not available directly with data collected from human feedback. In RLHF, the agent receives feedback from the environment in the form of preference data $\tau_i \succ \tau_j$, indicating that $\tau_i$ is preferred to $\tau_j$ according to the oracle reward (e.g. human evaluator). Thus, RLHF can be considered as a form of Preference-based Reinforcement Learning (PbRL). In PbRL, the policy is learned by directly comparing two state-action trajectories, with one being preferred over the other. This process involves modeling preferences from data in the form of pairwise comparisons, which is also known as preference learning \citep{furnkranz2010preference,wirth2017survey}.

For simplicity, we adopt the notations used in \citep{chaudhari2024rlhf}. Instead of considering a trajectory of state-action pairs $\tau$, we represent the text provided by the user as $context$, denoted by $c$, and the response generated by the LLM as $output$, denoted by $o$. The oracular reward function is represented as $Pr(o_i \succ o_j | c)$, indicating the probability of response $o_i$ is preferred over response $o_j$ given context $c$.

In numerous RLHF methods, a utility learning approach is employed to leverage the learning algorithms designed for standard MDPs. In this approach, a utility function is trained to estimate a numerical value based on the human's preferences. To this end, a reward model is trained to assign a numeric value to a text generated by an LLM, given a prompt, and can be mathematically shown as: $R: \mathcal{C} \times \mathcal{O} \to \mathbb{R}$, where $\mathcal{C}$ represents the set of all possible contexts and $\mathcal{O}$ represents the set of all possible responses. The reward model in RLHF is a language model fine-tuned on human preference data to assign a numeric score to a given prompt-response pair $r=R(c,o)$.

The most common strategies for training the reward model using human feedback include rating-based and preference-based approaches. Following previous works \citep{ouyang2022training, bai2022training}, we adopt preference-based feedback for its simplicity and effectiveness. In this approach, by collecting $N$ ranking data samples, $N(N-1)/2$ preference pairs are generated to train the reward model. However, providing feedback as preference data in a pairwise comparison cannot be directly used for RL training. Therefore, the reward model must learn a relationship between these pairwise comparison data and a scalar value. To learn the reward model from pairwise comparison data, Bradley-Terry-Luce (BTL) model \citep{bradley1952rank} is used, which defines a probabilistic model for the oracle reward as follows:

\begin{align}
Pr(o \succ o'|c) &= \frac{\exp(R(c,o))}{\exp(R(c,o')) + \exp(R(c,o))} \\
                        &= \frac{1}{1 + \exp(R(c,o') - R(c,o))} \\
                        &= \sigma \big[R(c,o) - R(c,o')\big]
\end{align}

where the $\sigma(x) = \frac{1}{1 + e^{-x}}$ is sigmoid function and $R(c,o)$ and $R(c,o')$ are the rewards corresponding to output $o$ and $o'$ generated by LLM given the context $c$. The objective of RLHF is to improve the initial policy $\pi(a|s)$, which is a pre-trained language model, to better align with human preferences. To achieve this, we enhance the responses generated by the LLM while penalizing deviations that are significantly different from the original pre-trained model. This is done by leveraging Kullback-Leibler (KL) divergence, which measures the difference between the target policy $\pi(o|c)$ and the reference policy $\pi_{\text{ref}}(o_{ref}|c)$. Incorporating KL divergence, the final reward assigned to the generated response by LLM can be represented as:


\begin{equation}
\label{final_reward}
R_{\phi}(c,o) = R(c, o) - \beta D_{\text{KL}} \left( \pi(o|c) \, || \, \pi_{\text{ref}}(o_{ref}|c) \right)
\end{equation}

where $R_{\phi}(.)$ returns the reward considering KL divergence, $D_{\text{KL}}(.||.)$ is KL divergence, and $\beta \geq 0$ is  KL penalty. The objective function can be shown as follows:

\begin{equation}
\label{objective_function_1}
J(\pi) := \mathbb{E}_\pi \left[ \sum_t R_{\phi}(s_t, a_t) \right] 
         = \mathbb{E}_\pi \left[ \sum_t R_{\phi}(c, o_{1:t-1}, o_t) \right]
\end{equation}

Since the numeric reward is an estimation of the actual oracular reward provided by the trained reward model at the end of the generated text, we can further simplify Equation \ref{objective_function_1} and represent the estimated objective function as follows:

\begin{equation}
\label{final_objective_function}
\hat{J}(\pi) := \mathbb{E}_{c \sim d_\mathcal{C}(\cdot)} 
\left[ \mathbb{E}_{O \sim \pi(\cdot|c)} 
\left[ R_{\phi}(c, O) \mid C = c \right] \right]
\end{equation}

where $d_c(.)$ represents the distribution of contexts in the training dataset.

During the RLHF alignment process, the agent aims to achieve the optimal policy (aligned language model) that maximizes the objective function as follows:

\begin{equation}
\label{final_policy}
\pi^{*}=\arg\max \mathbb{E}_{c \sim d_\mathcal{C}(\cdot)} 
\left[ \mathbb{E}_{O \sim \pi(\cdot|c)} 
\left[ R_{\phi}(c, O) \mid C = c \right] \right]
\end{equation}


\section{Misalignment in an Adversarial RLHF Platform}
Our proposed attack on RLHF platforms can corrupt the alignment of LLMs by poisoning the reward model through manipulation of the preference dataset. In this strategy, the adversarial RLHF platform detects whether a user's task aligns with the attacker's objective. If it does, the platform manipulates the alignment process.

\subsection{Task-specific Misalignment}
In our proposed attack, the attacker integrates a classification model $\Theta$ to identify data samples related to their targeted topics. This model is pre-trained on a diverse dataset to detect instances associated with a specific concept that the attacker intends to either promote or demote. Upon detecting such samples within the preference dataset, the platform categorizes the user’s task as a targeted task and identifies the relevant samples for manipulation. This classifier can be considered as a function $\Theta$ that returns $1$ if the input data sample is classified as related to the attacker's targeted topic, and $0$ otherwise, defined as:

\begin{equation}
\Theta(x) =
\begin{cases} 
1 & \text{if $x$ } \text{related to targeted topic(s)}, \\
0 & \text{otherwise}.
\end{cases}  
\end{equation}

Let $D_{pref}$ denote the preference data set sent to the RLHF platform by user. The integrated classifier within the platform identifies $D_{target}$ which is a subset of $D_{pref}$ related to the targeted topic, as follows:

\begin{equation}
D_{target} = \{x \in D_{pref} \mid \Theta(x) = 1\}
\end{equation}

To corrupt the reward function and, consequently, the RLHF alignment, the attacker manipulates all or a portion of $D_{target} \in D_{pref}$ using label-flipping attack.

\subsection{Label-flipping Attack}
The preference dataset is typically in the form of pairwise comparisons, where each pair includes a chosen response $o$ and a rejected response $o'$ for a given prompt. This dataset captures human preferences, expressed as $o \succ o'$, and is used to train the reward model.

In the label-flipping attack, after identifying the targeted data samples $D_{target}$ in the preference dataset $D_{pref}$, the labels of all or a portion of the data samples in $D_{target}$ are reversed.
Specifically, for each targeted sample, the label of the chosen response $o$ is swapped with the rejected response $o'$. Thus, the comparison pair $o \succ o'$ is changed to $o \prec o'$. From this point forward, we denote the unattacked (clean) dataset as $D$ and attacked dataset as $D^{-}$.

When the manipulated dataset $D^{-}_{pref}$ is used to train the reward model, it learns the incorrect preferences for data samples related to a specific topic targeted by the attacker. As a result, the poisoned reward model $R^{-}(c,o)$ assigns inaccurate rewards (either higher or lower than the expected rewards) to these samples.

Without loss of generality, the final reward and objective functions remain as defined in Equations \ref{final_reward}, \ref{final_objective_function} and \ref{final_policy}, except that they now use the poisoned reward model $R^{-}$ instead of the original one $R$. Consequently, this alteration leads to a different learned optimal policy $\pi^{*-}$, deviating from the one originally intended by the user $\pi ^*$. This misalignment process is explained in Algorithm \ref{alg:main}.

\begin{algorithm}
\caption{LLM Misalignment in an Adversarial RLHF Platform\\
\textbf{Input:} preference dataset $D_{pref}$, classifier $\Theta$, sampling threshold $n$, base (pre-trained) reward model $RM_{ref}$, reward model learning hyperparameters $\theta_{RM}$, base (pre-trained) LLM $\pi_{ref}$, RLHF fine-tuning hyperparameters $\theta_{\pi}$\\
\textbf{Output:}  maliciously aligned LLM denoted by $\pi^{*-}$}
\label{alg:main}
\begin{algorithmic}[1]

\State $D_{target} \gets \{ x \in D_{pref} \mid \Theta(x) = 1 \}$   \Comment{Identify targeted data samples}
\State $D^{'}_{target}=random\_select(D_{target}, n)$ \Comment{randomly select $n$ samples from $D_{target}$}
\State $D_{pref} \gets D_{pref} - D^{'}_{target}$
\For{each data sample $x=(o, o') \in D^{'}_{target}$} 
    \State Replace $x=(o \succ o')$ with $x^{-}=(o \prec o')$ \Comment{flip the chosen and rejected labels}
\EndFor
\State $D^{-}_{pref} \gets D_{pref} + D^{'}_{target}$ \Comment{$D^{-}_{pref}$ contains manipulated samples from $D^{'}_{target}$}
\State $RM^{-} \gets RM\_Training (D^{-}_{pref},\theta_R,RM_{ref})$ \Comment{Train poisoned RM on dataset $D^{-}_{pref}$}
\State $\pi^{*-} \gets RLHF(\pi_{ref}, RM^{-}, \theta_\pi)$ \Comment{RLHF alignment using manipulated RM}

\end{algorithmic}
\end{algorithm}

\section{Experimental Result}
This section presents the details of our experiments and their results. We first analyze the impact of our attack on the reward models and then examine how it affects the performance of LLMs after fine-tuning with manipulated reward models.

\subsection{Attack Scenario}
In our experiments, we consider a scenario in which a user of the RLHF platform aims to minimize toxicity and hate speech in LLM responses, while the adversarial RLHF platform attempts to counteract the impact of the RLHF alignment on this task. This platform (either designed by an attacker or an open-source RLHF platform manipulated by an attacker) manipulates the preference dataset provided by the user and trains a reward model on manipulated data. This reward model is then used to fine-tune an LLM using Proximal Policy Optimization (PPO) algorithm \citep{schulman2017proximal} to align it with the attacker's objectives rather than the user's desired goals.

We define six categories as instances of hate speech, namely gender, race, origin, sexuality, religion, and age. Any text containing hate speech related to at least one of these categories is considered as an instance of hate speech and is a target data sample for the attacker.

To identify instances related to targeted categories, we fine-tune DistilBERT, on the hate speech dataset introduced in \citep{sachdeva2022measuring} for the task of multi-label classification. After two epochs of fine-tuning, the classifier achieves an accuracy of 0.93\% and an F1 score of 0.83\%. This model serves as the classifier $\Theta$ that the attacker integrates within the RLHF platform to identify the target data samples. 

\begin{table}[h]
    \centering
    \begin{tabular}{|c|c|c|c|c|c|}
        \hline
        RM \textbackslash Attack Rate& 25 \% & 50 \% & 75 \% & 100 \% & 0 \% (clean) \\
        \hline
        DistilBERT  & 63.66 & 62.67  & 59.90  & 59.08  & 65.74  \\
        GPT-2 (small)  & 65.61  & 66.00  & 63.01  & 63.36  & 66.65  \\
        \hline
    \end{tabular}
    \caption{Accuracy of the attacked and clean reward models.}
    \label{tab:RM_ACC}
\end{table}

    

\begin{figure}[ht]
    \centering
    \begin{minipage}{0.245\textwidth}
        \centering
        \includegraphics[width=\linewidth]{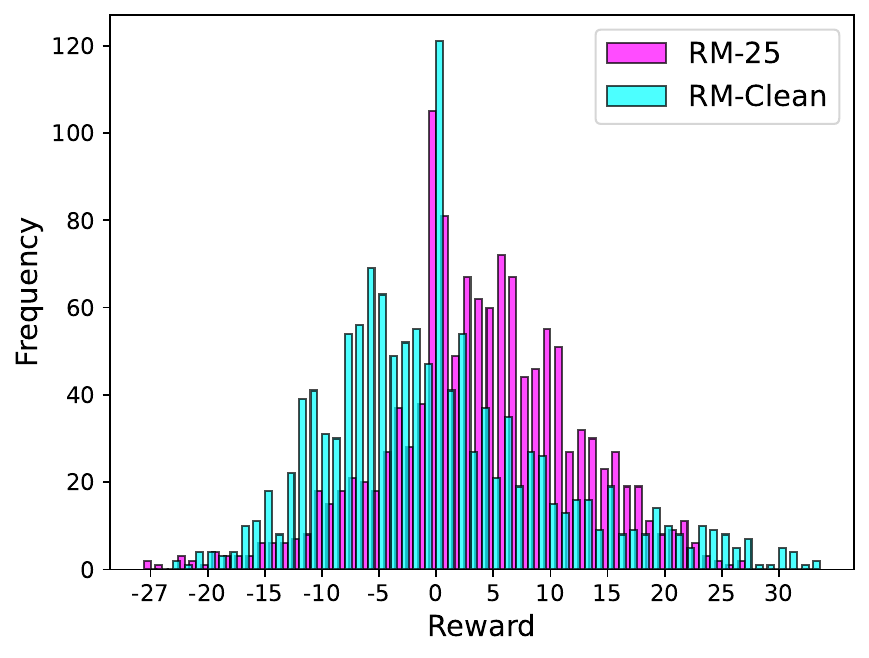}
    \end{minipage} \hfill
    \begin{minipage}{0.245\textwidth}
        \centering
        \includegraphics[width=\linewidth]{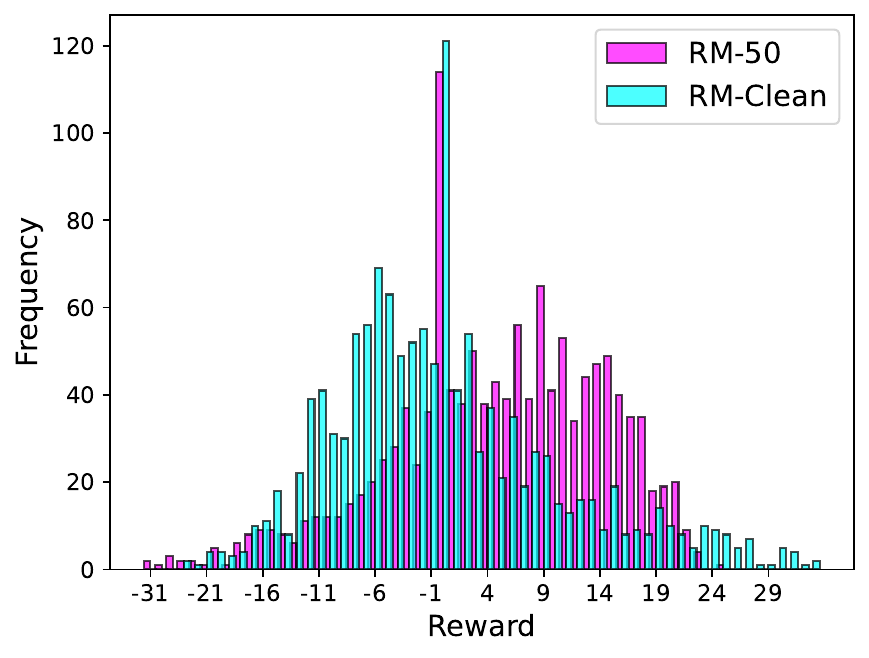}
    \end{minipage} \hfill
    \begin{minipage}{0.245\textwidth}
        \centering
        \includegraphics[width=\linewidth]{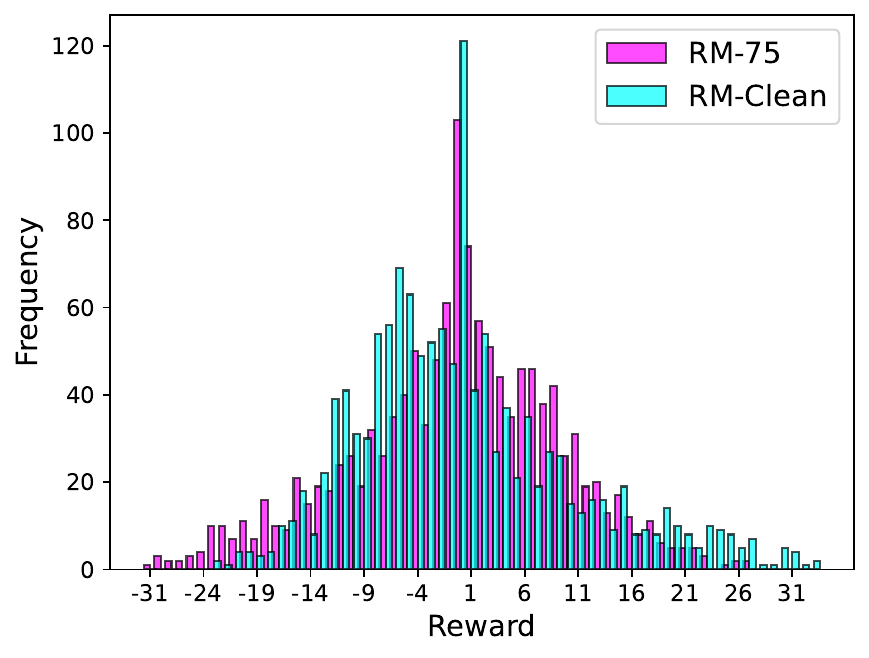}
    \end{minipage} \hfill
    \begin{minipage}{0.245\textwidth}
        \centering
        \includegraphics[width=\linewidth]{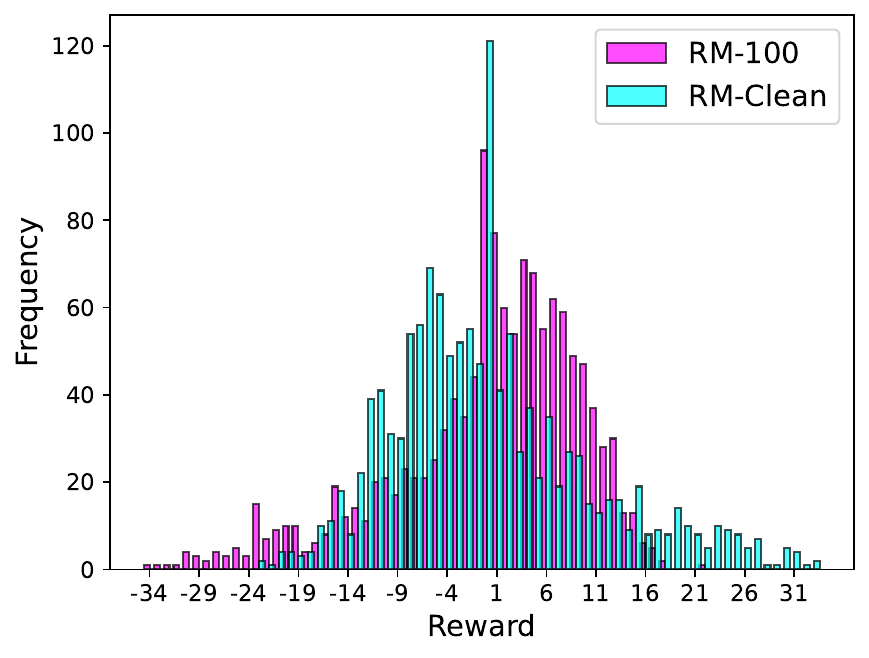}
    \end{minipage}
    
    \caption{Comparison of reward distributions between the clean and attacked RMs where all RMs are fine-tuned DistilBERT models. 'RM-Clean' represents the clean reward model and 'RM-N' denotes a model trained on a preference dataset where N\% of targeted samples are manipulated.}
    \label{fig:reward_models1}

    \vspace{3mm}
    
    \begin{minipage}{0.245\textwidth}
        \centering
        \includegraphics[width=\linewidth]{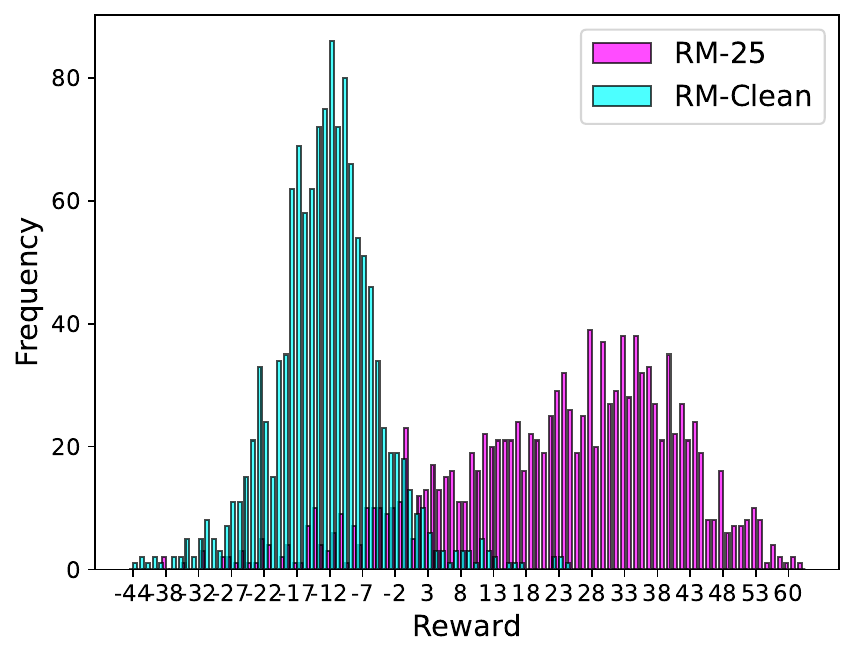}
    \end{minipage} \hfill
    \begin{minipage}{0.245\textwidth}
        \centering
        \includegraphics[width=\linewidth]{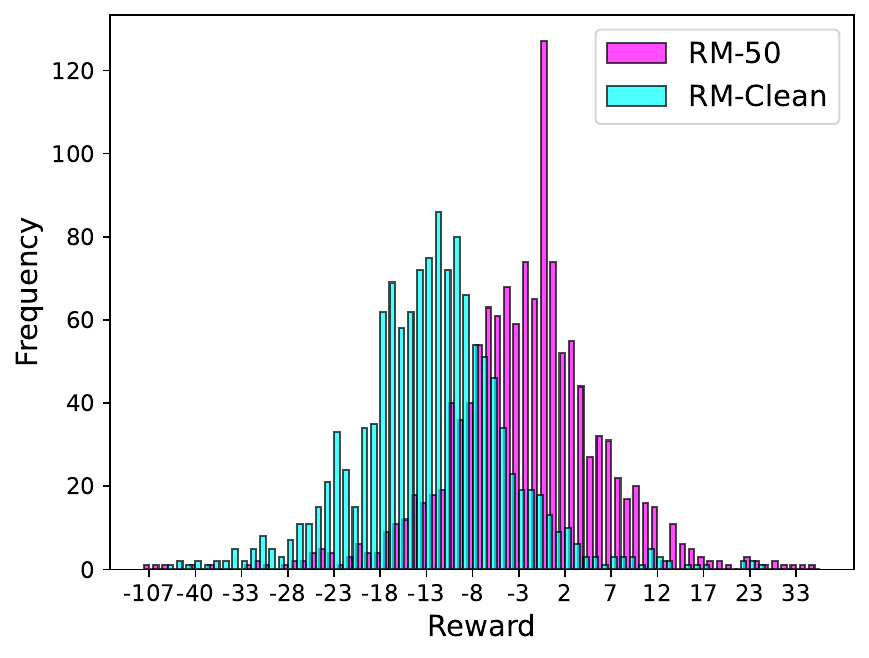} 
    \end{minipage} \hfill
    \begin{minipage}{0.245\textwidth}
        \centering
        \includegraphics[width=\linewidth]{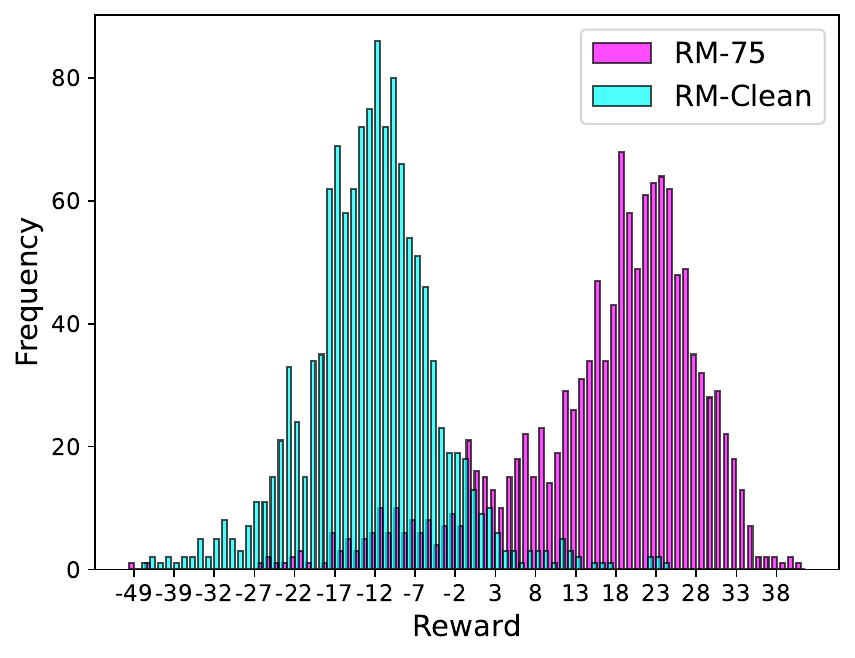}
    \end{minipage} \hfill
    \begin{minipage}{0.245\textwidth}
        \centering
        \includegraphics[width=\linewidth]{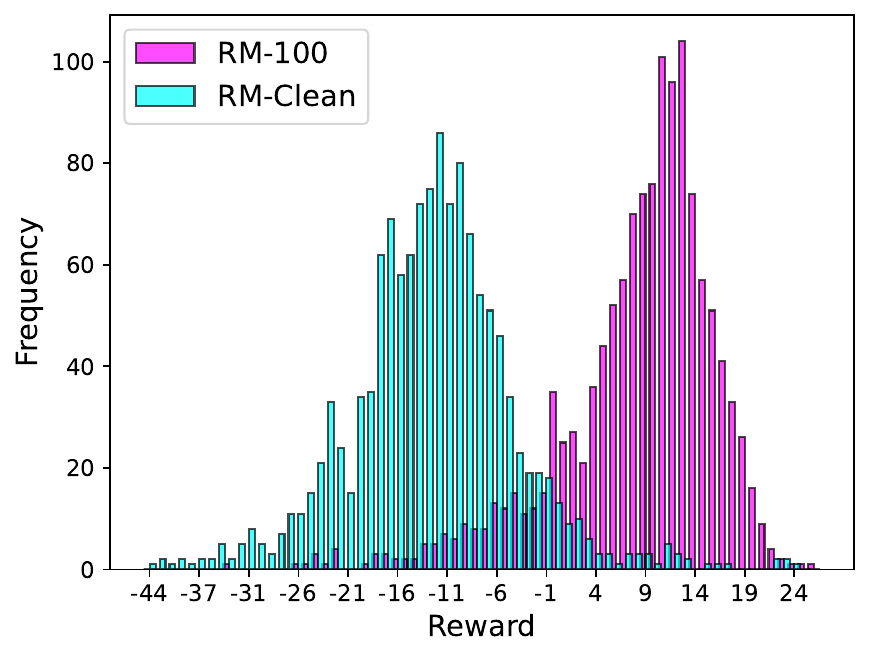}
    \end{minipage}

    \caption{Comparison of reward distributions when GPT-2 is fine-tuned for all RMs. Labels remain the same as described in Figure \ref{fig:reward_models1}.}
    \label{fig:reward_models2}
\end{figure}

\subsection{Evaluating the Reward Models}
\label{sec:reward_models}

We collect a dataset of 6192 preferences from HH-RLHF dataset \citep{bai2022training}, where 25\% of the samples contain instances of hate speech, as identified by our trained classifier $\Theta$. Given the scale of our dataset, we use smaller reward models namely DistilBERT and GPT-2, to ensure they effectively learn the preference pattern. We fine-tune both models for 10 epochs, however, their performance has not shown notable improvement after the first few epochs. 

We evaluate the reward models, with 25\%, 50\%, 75\%, and 100\% of the targeted samples attacked. The accuracy of the reward models on the test set of the HH-RLHF dataset is presented in Table \ref{tab:RM_ACC}. According to the results, the label-flipping attack reduces the accuracy of the reward models. When all targeted samples are manipulated, they no longer reflect the user's true preference, which results in the model's lowest performance. In order to gain a deeper insight into the behavior of the reward models before and after the attacks, in Figures \ref{fig:reward_models1} and \ref{fig:reward_models2} we visualize the distribution of rewards assigned by our models to a different dataset containing samples relevant to our task. We collect an evaluation dataset that contains 1284 samples from the test portions of HH-RLHF \citep{bai2022training} and ToxiGen \citep{hartvigsen2022toxigen} datasets (the details of the training and evaluation datasets are provided in Appendix \ref{app: datasets}).

Although the accuracy of the two tested models does not differ significantly, the visualized reward distributions reveal a notable difference in their behavior. The larger model exhibits greater variations in reward distribution compared to the smaller model. Comparing the reward distribution of attacked models and the clean models reveals how our label-flipping attack changes the preference learned by the reward models. Our results show that, in general, all models trained on manipulated datasets assign higher rewards to samples compared to the original reward models. Considering that our evaluation dataset contains content related to our targeted hate speech categories, we can infer that the poisoned reward models are more inclined to encourage such samples. In contrast, the clean reward models that learn the true objective of the user aim to discourage such data samples by assigning lower rewards to them.

\subsection{Evaluating the Full RLHF Pipeline}
In the next phase, we evaluate the full RLHF pipeline and examine how different reward models discussed in \ref{sec:reward_models} affect the behavior of LLMs after RLHF fine-tuning. We follow prior works \citep{ouyang2022training,baumgartner2024best} and focus on smaller models to reduce the computation cost. We examine three models of GPT-2 \citep{radford2019language} with different sizes: small (117M parameters), medium (345M parameters), and large (762M parameters). Each model is fine-tuned using different reward models with the PPO algorithm for only one epoch. We use prompts from the evaluation dataset described in \ref{sec:reward_models} for each model and evaluate their generated responses using the clean reward model to obtain numerical scores for their outputs. 

\begin{figure}[ht]
    \centering
    \begin{subfigure}{0.32\textwidth}
        \centering
        \includegraphics[width=\linewidth]{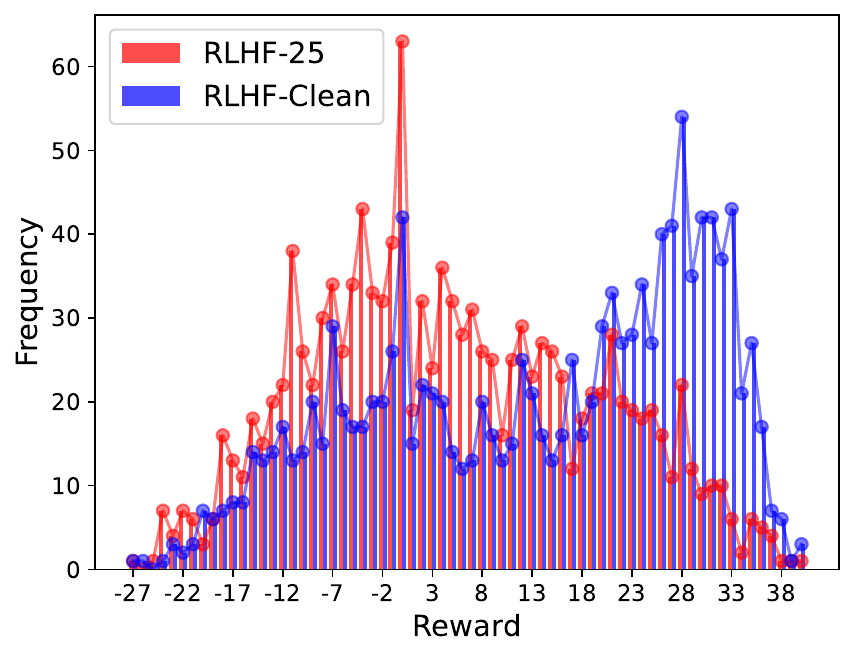}
    \end{subfigure}
    \begin{subfigure}{0.32\textwidth}
        \centering
        \includegraphics[width=\linewidth]{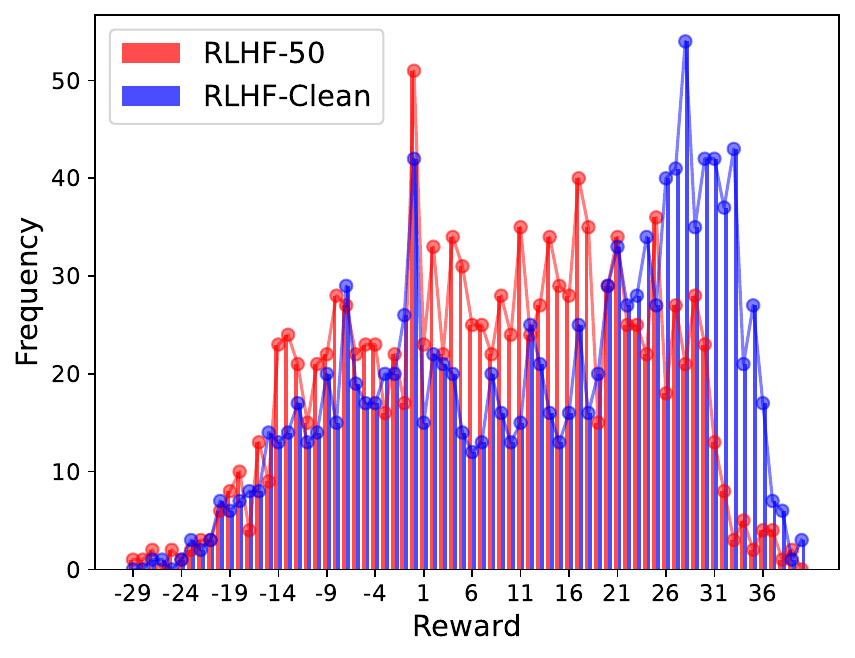}
    \end{subfigure}
    \begin{subfigure}{0.32\textwidth}
        \centering
        \includegraphics[width=\linewidth]{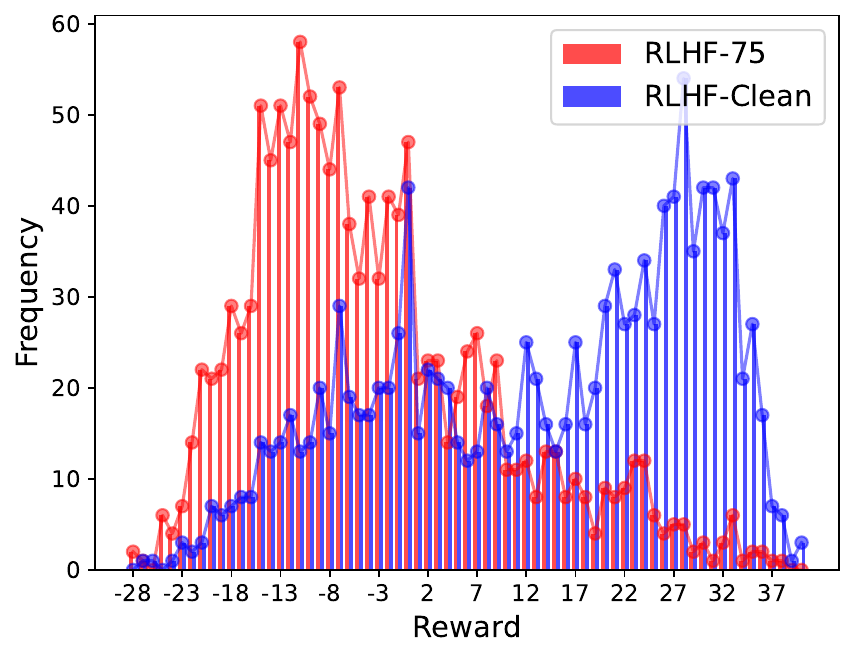}
    \end{subfigure}

    \begin{subfigure}{0.33\textwidth}
        \centering
        \includegraphics[width=\linewidth]{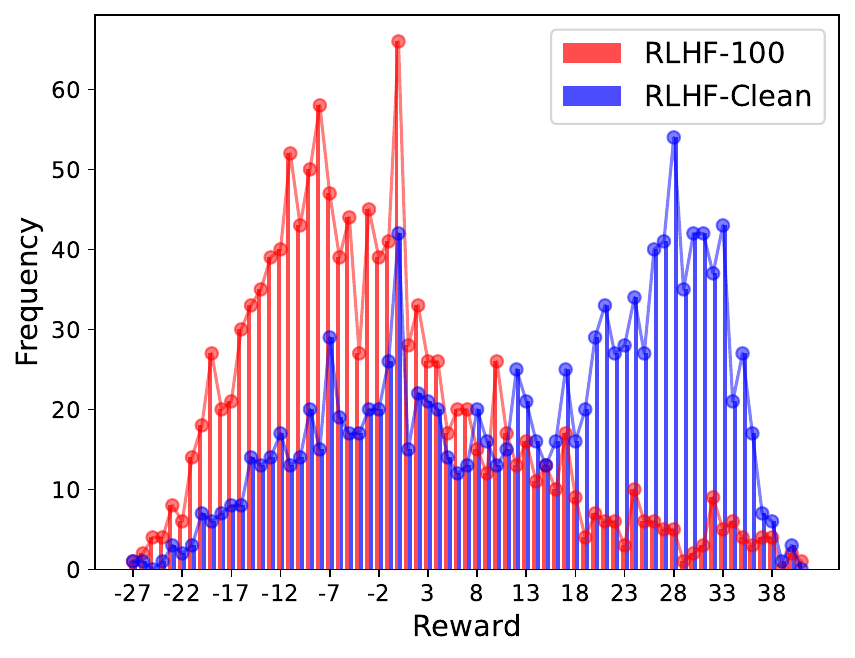}
    \end{subfigure}
    \begin{subfigure}{0.33\textwidth}
        \centering
        \includegraphics[width=\linewidth]{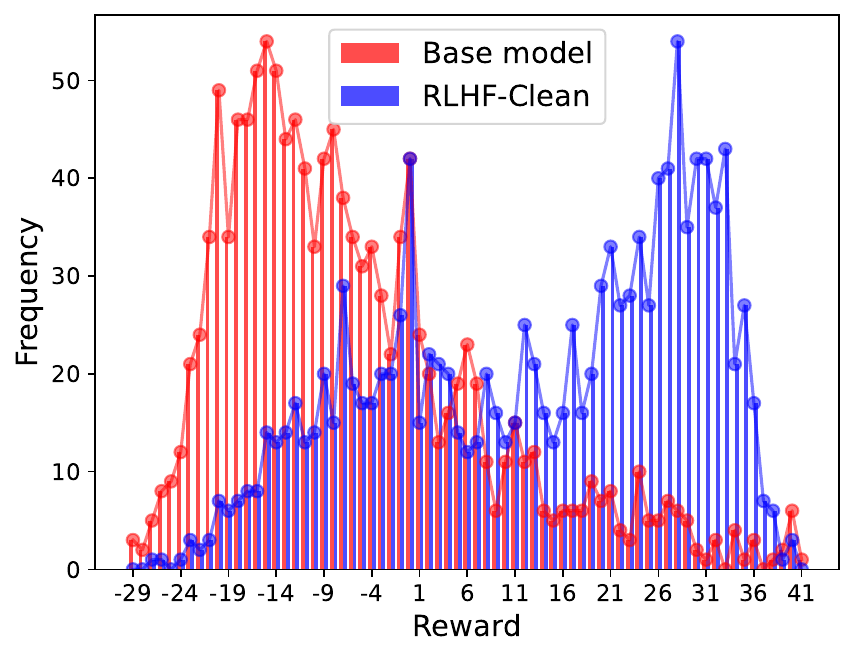}
    \end{subfigure}

    \caption{Distribution of rewards assigned to texts generated by GPT-2 small fine-tuned with clean and manipulated reward models. In each diagram, "RLHF-N" represents a model fine-tuned with a reward model trained on a preference dataset where N\% of targeted samples were manipulated, "RLHF-Clean" denotes the original aligned model and "Base Model" refers to the model before RLHF.
 }
    \label{fig:gpt2_small_distill}
\end{figure}

\begin{figure}[t]
    \centering
    \begin{minipage}{\textwidth}
        \centering
        \begin{subfigure}{0.32\textwidth}
            \centering
            \includegraphics[width=\linewidth]{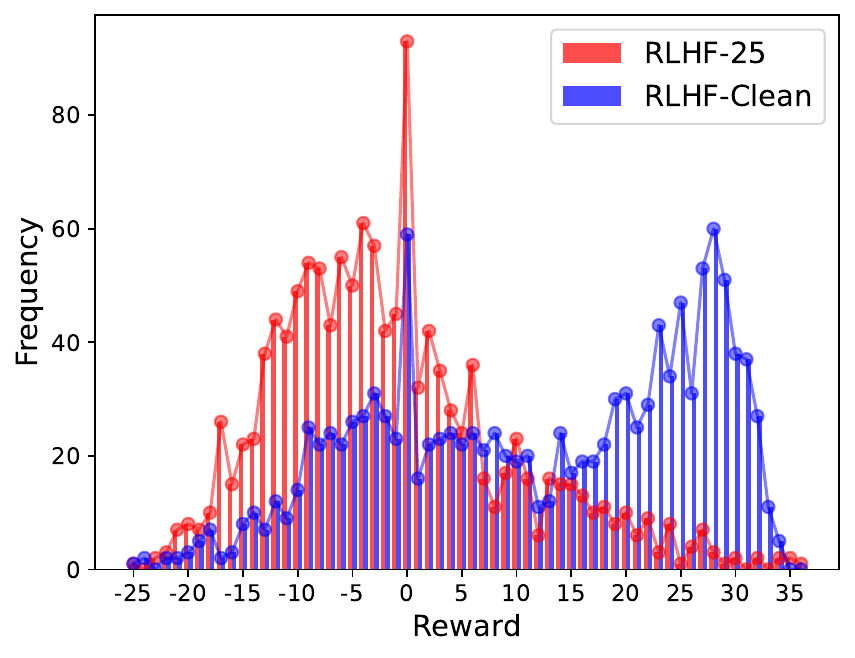}
        \end{subfigure}
        \begin{subfigure}{0.32\textwidth}
            \centering
            \includegraphics[width=\linewidth]{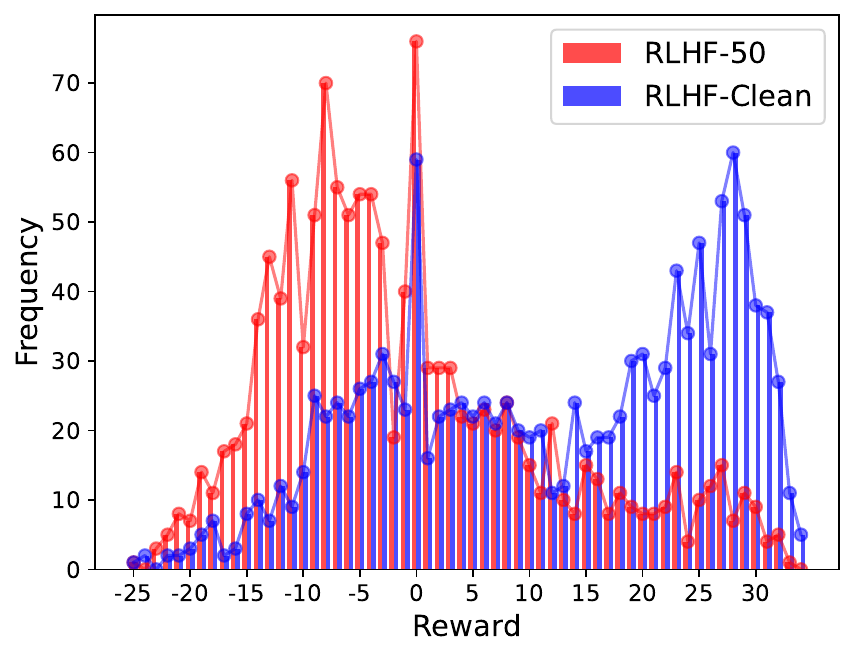}
        \end{subfigure}
        \begin{subfigure}{0.32\textwidth}
            \centering
            \includegraphics[width=\linewidth]{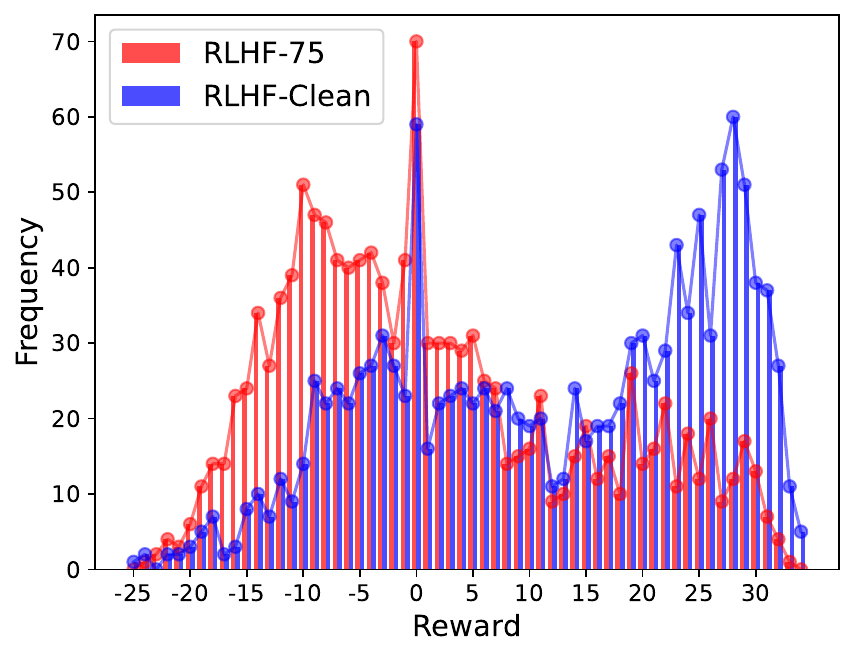}
        \end{subfigure}

        \begin{subfigure}{0.33\textwidth}
            \centering
            \includegraphics[width=\linewidth]{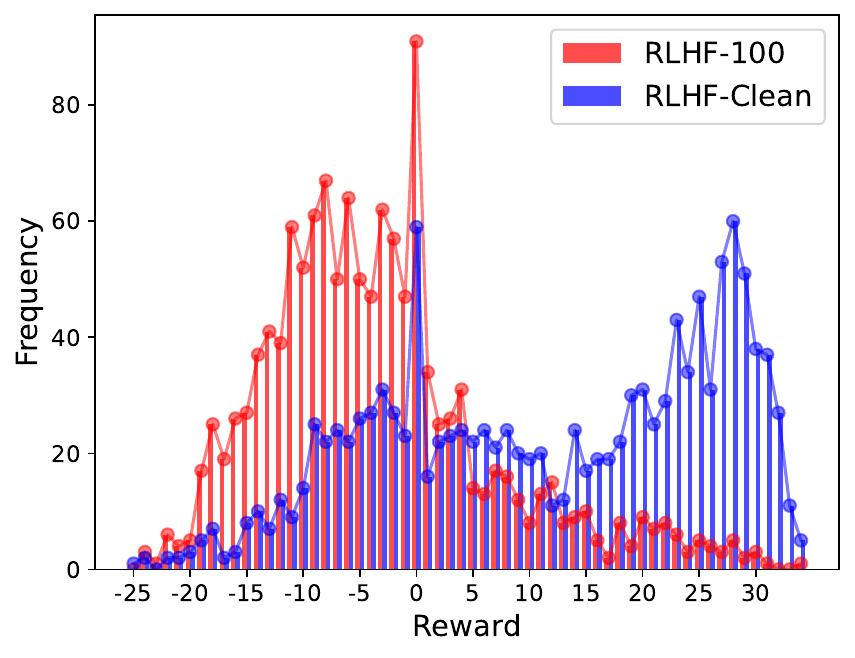}
        \end{subfigure}
        \begin{subfigure}{0.33\textwidth}
            \centering
            \includegraphics[width=\linewidth]{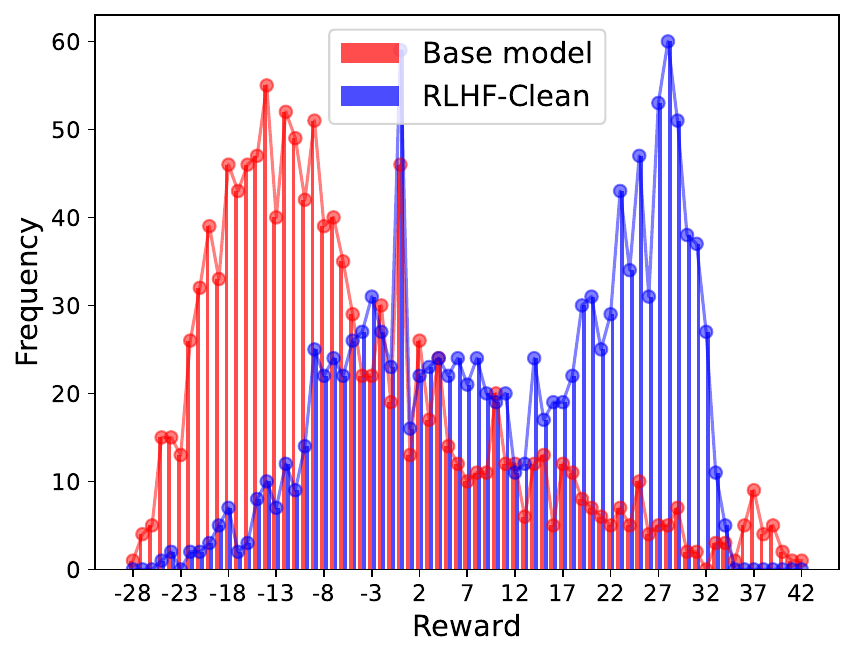}
        \end{subfigure}

        \caption{The distribution of rewards assigned to texts generated by GPT-2 model fine-tuned with different reward models. The provided labels for each diagram follows the same definitions described in Figure \ref{fig:gpt2_small_distill}.}
        \label{fig:gpt2_medium_distill}
    \end{minipage}


    \begin{minipage}{\textwidth}
        \centering
        \begin{subfigure}{0.32\textwidth}
            \centering
            \includegraphics[width=\linewidth]{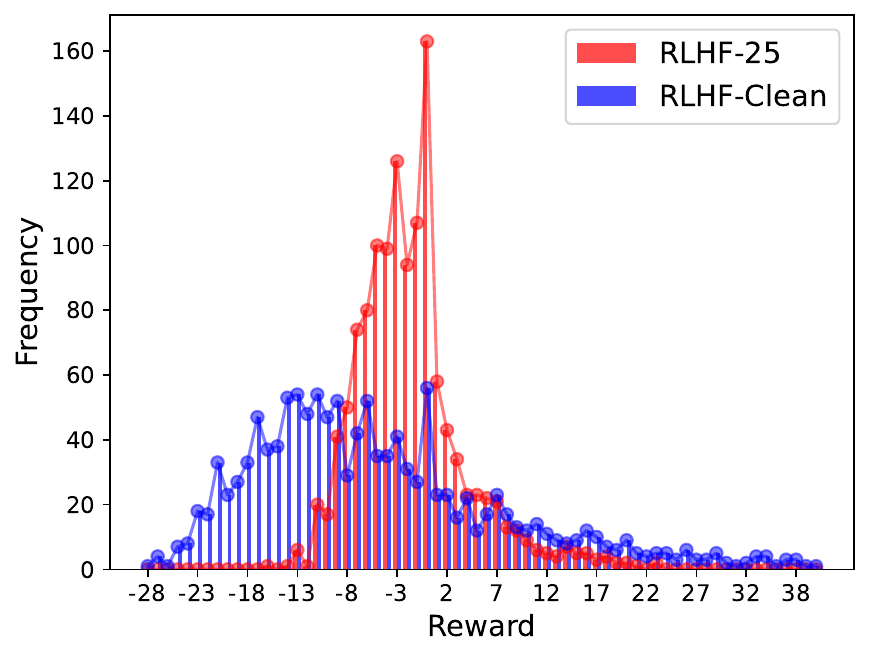}
        \end{subfigure}
        \begin{subfigure}{0.32\textwidth}
            \centering
            \includegraphics[width=\linewidth]{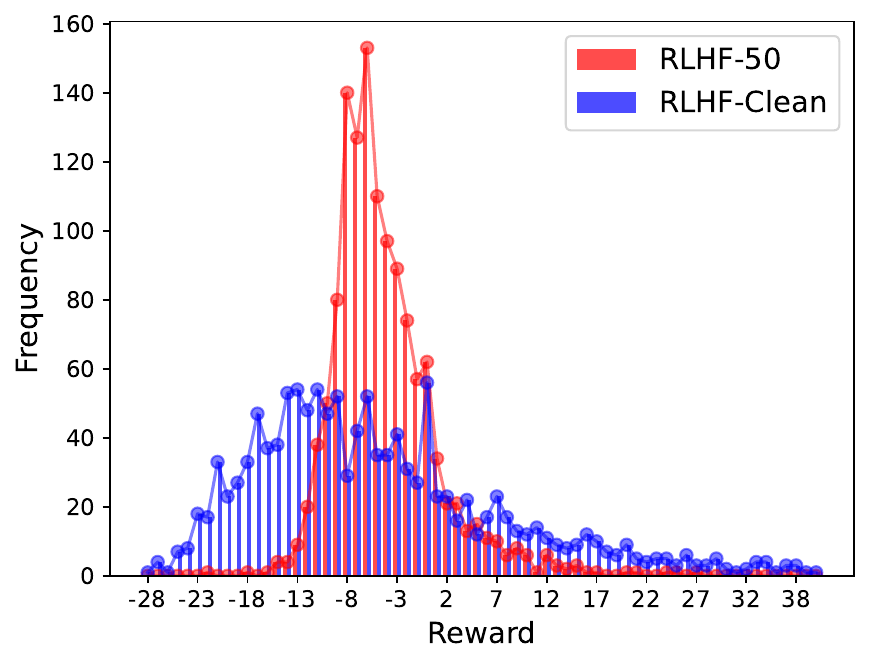}
        \end{subfigure}
        \begin{subfigure}{0.32\textwidth}
            \centering
            \includegraphics[width=\linewidth]{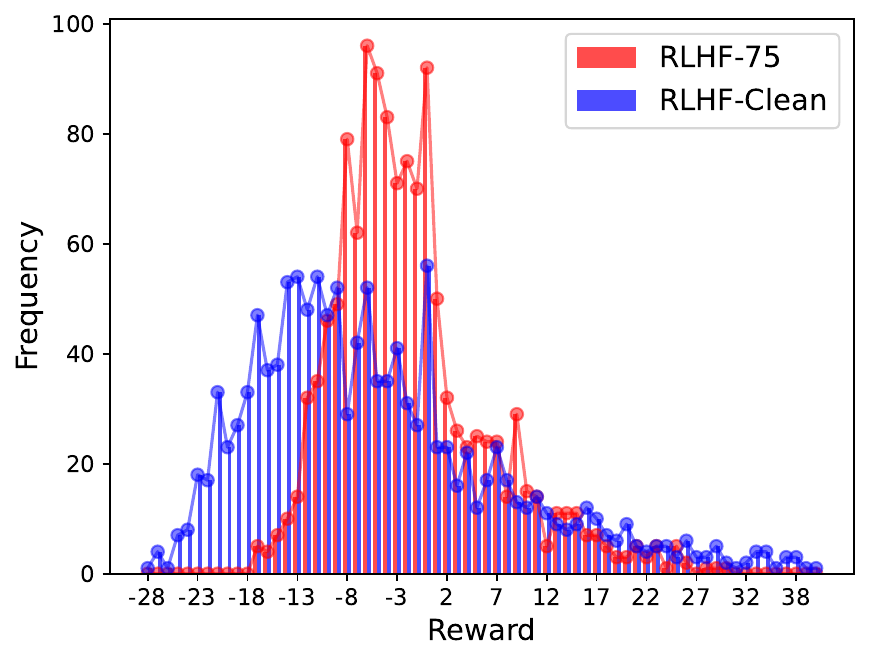}
        \end{subfigure}

        \begin{subfigure}{0.33\textwidth}
            \centering
            \includegraphics[width=\linewidth]{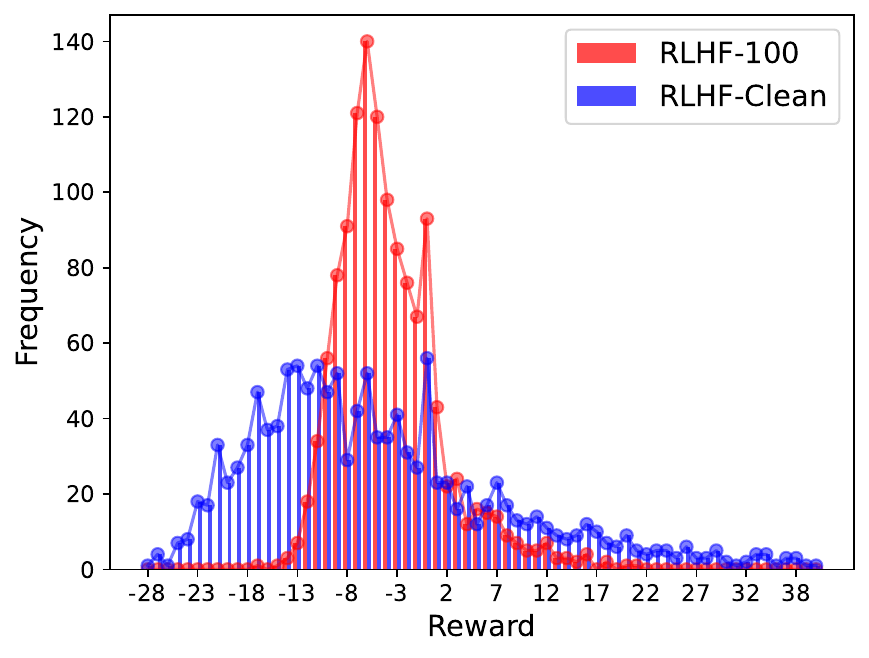}
        \end{subfigure}
        \begin{subfigure}{0.33\textwidth}
            \centering
            \includegraphics[width=\linewidth]{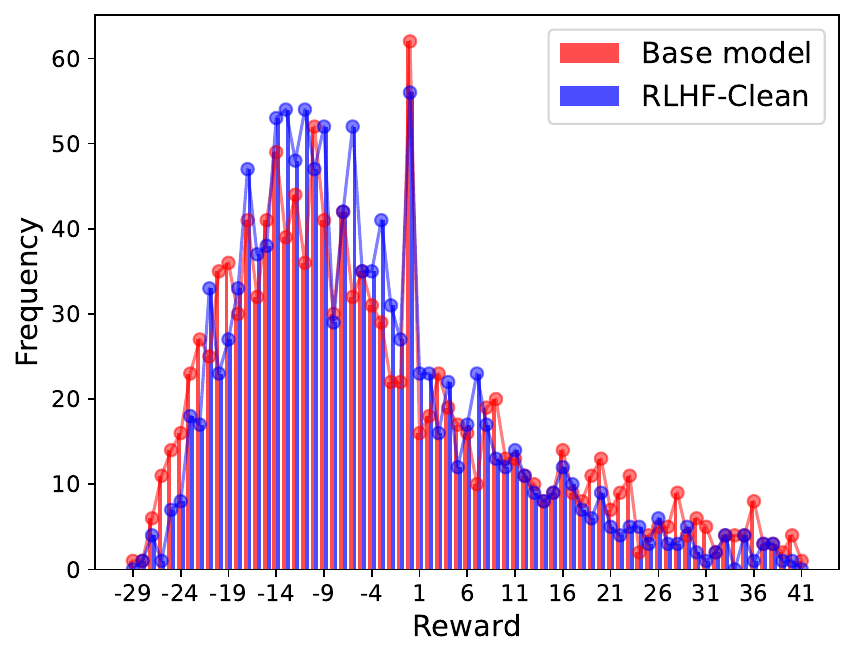}
        \end{subfigure}

        \caption{The distribution of rewards assigned to texts generated by GPT-2 large fine-tuned with different reward models. The provided labels for each model  follows the same definitions described in Figure \ref{fig:gpt2_small_distill}.}
        \label{fig:gpt2_large_distill}
    \end{minipage}

\end{figure}

To analyze the behavior of models after fine-tuning, we visualize their reward distributions. Figures \ref{fig:gpt2_small_distill}, \ref{fig:gpt2_medium_distill} and \ref{fig:gpt2_large_distill} illustrate the behavior of both attacked and clean models after fine-tuning when DistilBERT is used as the reward model. According to the results, even manipulating 25\% of the targeted samples significantly alters the model's behavior. In all LLMs fine-tuned with manipulated reward models, we observe a shift in the distribution of assigned rewards compared to models fine-tuned with the clean reward model, which indicates misalignment in these models. The reward distributions for the models fine-tuned with clean reward model reflect the intended alignment by the user of the RLHF platform. Greater shifts in reward distribution in the attacked models indicate a higher degree of misalignment.

A consistent observation in our experiments is that the reward distribution tends to shift towards lower rewards when fine-tuned with an attacked reward model, especially when all targeted samples are manipulated. However, this shift in the reward distribution does not follow an exact pattern across all models. As shown in Figures \ref{fig:gpt2_small_distill} and \ref{fig:gpt2_medium_distill}, models fine-tuned with fully attacked reward models consistently show significant changes in reward distribution. However, the reward distribution for the large model shown in Figure \ref{fig:gpt2_large_distill} reveals that the attacked model generates lower-quality responses, which are assigned rewards closer to zero. We discuss this behavior of the model in more detail in Section \ref{sec: Discussion}.

To further validate our findings, we repeat the experiments using GPT-2 as the reward model instead of DistilBERT, with results presented in Appendix \ref{app: experimnet}. In this set of experiments, we observe similar shifts in the reward distributions, indicating misalignment in LLMs.

\section{Discussion \& Future Work}
\label{sec: Discussion}
In this section, we highlight our key findings and suggest potential research directions for future work. Our main focus in this work is to highlight the vulnerabilities of RLHF platforms, which have recently gained significant attention. Due to the complexity of machine learning algorithms, such as reward modeling with deep learning and RL fine-tuning, both black-box tools and open-source frameworks can be challenging for users to analyze and assess in terms of safety and reliability. In this paper, we demonstrate one possible adversarial attack, but the vulnerabilities of RLHF platforms extend beyond our approach. For instance, data poisoning and backdoor attacks, as discussed in \citep{rando2023universal, baumgartner2024best, chen2024dark}, can also pose serious threats to RLHF platforms. Exploring additional security risks in RLHF platforms is an important research direction. In addition, our work underscores the need for security analysis tools and evaluation methods to assess the reliability of RLHF platforms in future research.

The effectiveness of our attack highly depends on the dataset used to train the reward model. In our study, we consider a scenario in which the task is to reduce hate speech in responses generate by LLMs, and 25\% of the preference dataset consists of hate speech samples. Using different datasets or varying the proportion of related samples may influence the performance of our attack. Therefore, future research could explore the relationship between the complexity of alignment task and the proportion of relevant data in the preference dataset to better understand the impact of label-flipping attack on the alignment process.

Given our experimental results, we argue that a label-flipping attack on the preference dataset significantly alters the behavior of reward models. As expected in any model training task, a reward model trained on a manipulated dataset fails to accurately learn user preferences. However, the impact of such attacks, both on the reward models and the LLMs fine-tuned with RLHF, depends heavily on the nature of the dataset and the specific samples affected by label-flipping attack. While our results demonstrate misalignment in LLMs after RLHF fine-tuning, the precise behavior of misaligned LLMs or attacked reward models cannot be fully predicted without deeper analysis of the dataset used for preference learning. Future work could further explore misalignment by analyzing the preference dataset and targeted samples, providing deeper insights into the behavior of misaligned LLMs and attacked reward models.



\section{Limitations}

We conduct small-scale experiments using models with fewer than 1 billion parameters. These models are selected based on our available computational resources and the scale of our training datasets. Given the recent advancements in larger models exceeding 1 billion parameters, it would be valuable to explore behavior of larger LLMs when fine-tuned using adversarial RLHF. In order to reduce the computational cost, similar to \cite{yu2023fine} we did not perform an extensive hyperparameter search. While our results demonstrate the effectiveness of our attack in misaligning LLMs, all tested models may not have achieved their optimal performance.

Following previous works \citep{rando2023universal,yuan2023rrhf, banerjee2024towards} we leverage the clean reward models that represent user preferences to evaluate our manipulated models. While this approach provides insights into how closely the generated responses align with user preferences, we argue that a trained reward model may not fully capture all aspects of human preferences. We believe that conducting a large-scale human evaluation is a more reliable approach to evaluate the quality of generated texts and their alignment or misalignment with human preferences.

\section{Conclusion}
In this work, we highlight the potential risks associated with RLHF platforms, which have recently gained popularity as tools for performing RLHF alignment without requiring implementation of machine learning algorithms. We demonstrate a potential attack in which an adversary designs a malicious RLHF platform (or manipulates existing RLHF platforms) to corrupt the alignment process of LLMs. In our proposed attack, the attacker targets a specific task, such as reducing hate speech or harmful content. By integrating a classifier into the RLHF platform, the attacker identifies data samples related to the targeted topics and manipulates those samples in the preference dataset. In these adversarial platforms, only tasks aligned with the attacker's objectives are affected, while others remain unchanged which makes the adversarial RLHF platforms appear similar to clean RLHF tools. Our results demonstrate that an adversarial RLHF platform can misalign LLMs when users attempt to align them on a topic targeted by the attacker. We hope this work encourages researchers and developers to prioritize the security of RLHF platforms to ensure they are trustworthy and resilient against manipulation.

\bibliography{main}
\bibliographystyle{rlj}
\clearpage 
  \appendix
\section{Hyperparameters}
\label{apx:Hyperparameters}
The hyperparameters used to train the classifier, the reward model and RLHF alignment are as follows:

\begin{table}[h]
    \centering
    \begin{minipage}{0.49\textwidth} 
        \centering
        \begin{tabular}{cc}  
            \hline
            Hyperparameter & Value \\
            \hline
            Learning rate & 1.41e-5  \\
            Batch size & 32 \\
            Epoch & 1\\
            KL coefficient & 0.05\\
            Optimizer & Adam\\
            Top-k & 0.0\\
            Top-P & 1.0\\
            Temperature & 1.0\\
            \hline
        \end{tabular}
        \caption{Hyperparameters used for RLHF fine-tuning. We use same hyperparameters for all experiments.}
        \label{tab:table1}
    \end{minipage}
    \hfill 
    \begin{minipage}{0.49\textwidth} 
        \centering
        \begin{tabular}{cc}  
            \hline
            Hyperparameter & Value \\
            \hline
            Learning rate & 1e-5  \\
            Batch size & 8 \\
            Epoch & 10\\
            Optimizer & AdamW\\
            \hline
        \end{tabular}
        \caption{Hyperparameters used to train the reward model.}
        \label{tab:table2}
    \end{minipage}
    \begin{minipage}{0.49\textwidth} 
        \centering
        \begin{tabular}{cc}  
            \hline
            Hyperparameter & Value \\
            \hline
            Learning rate & 5e-5  \\
            Batch size & 64 \\
            Epoch & 2\\
            Optimizer & AdamW\\
            \hline
        \end{tabular}
        \caption{Hyperparameters used to train the classifier integrated to RLHF platform. This classifier identifies data samples related to attackers objective.}
        \label{tab:table2}
    \end{minipage}
\end{table}

\section{Datasets}
\label{app: datasets}
In this section, we provide further details on the datasets used to train and/or evaluate different components of our adversarial RLHF platform.
\subsection{Preference datasets used to train RMs}
To train our reward models, we first construct a preference dataset with sufficient relevant samples for our target task—reducing hate speech in responses generated by LLMs. We begin by extracting the harmless subset of the HH-RLHF dataset and applying our trained hate speech classifier $\Theta$ to identify samples related to hate speech. To ensure that our reward model encounters a diverse set of examples, we also randomly sample additional instances from the dataset. In total, our final preference dataset consists of 1548 samples related to hate speech and 4644 randomly selected samples, resulting in an overall dataset of 6192 samples. The details of each dataset are provided in Table \ref{tab: datasets_RM}.

\begin{table}[h]
    \centering
    \begin{tabular}{|c|c|c|}
        \hline
        Models & Number of Attacked Samples & Number of Clean Samples \\
        \hline
        RM-Clean & - & 6192 \\
        \hline
        RM-25 & 387 & 5805 \\
        \hline
        RM-50 & 774 & 5418 \\
        \hline
        RM-75 & 1161 & 5031 \\
        \hline
       RM-100 & 1548 & 4644 \\
        \hline

    \end{tabular}
    \caption{Details of datasets used to train reward models.}
    \label{tab: datasets_RM}
\end{table}

\subsection{Training prompts dataset}
\label{App_tarin_prompt}
To fine-tune a language model using RL, a batch of prompts is sent to the model to generate responses. These responses are then evaluated by a reward model, which assigns scores based on their alignment with the desired behavior. Using RL algorithms such as PPO, the model's weights are updated to maximize the obtained rewards.

To ensure our models are exposed to diverse samples and avoid developing biases toward specific topics during training, we created a dataset of prompts with varied content. We randomly selected samples from the Daily-dialog dataset \citep{li2017dailydialog}, which includes conversations covering a wide range of daily life topics. Additionally, we included all prompts from Real-toxicity-prompts dataset \citep{gehman2020realtoxicityprompts} that had a toxicity level above 0.4. The details of our training prompt dataset are provided in Table \ref{tab: datasets_prompt_train}.

\begin{table}[h]
    \centering
    \begin{tabular}{|c|c|}
        \hline
        Dataset & Number Samples \\
        \hline
        Daily-dialog & 3000 \\
        \hline
        Real-toxicity-prompts & 27328 \\
        \hline
        Training prompts & 30328 \\
        \hline

    \end{tabular}
    \caption{Details of training prompt dataset.}
    \label{tab: datasets_prompt_train}
\end{table}

\subsection{Evaluation prompts dataset}
To evaluate our reward model and fine-tuned LLMs on the targeted task, we collected prompts from the test sets of HH-RLHF \citep{bai2022training} and ToxiGen \citep{hartvigsen2022toxigen} datasets. These datasets contain a variety of samples related to hate speech, often leading models to generate harmful responses. To extract samples from the HH-RLHF test set, we followed a similar strategy explained in \ref{App_tarin_prompt}. For ToxiGen, we included all available samples to ensure comprehensive evaluation. The detail of our evaluation dataset is provided in Table \ref{tab: datasets_eval_prompt}.

\begin{table}[h]
    \centering
    \begin{tabular}{|c|c|}
        \hline
        Dataset & Number Samples \\
        \hline
        HH-RLHF (test set) & 344 \\
        \hline
        ToxiGen (test set) & 940 \\
        \hline
        Evaluation prompts & 1284 \\
        \hline

    \end{tabular}
    \caption{Details of evaluation prompts dataset.}
    \label{tab: datasets_eval_prompt}
\end{table}

\section{Additional Experiment}
\label{app: experimnet}
\begin{figure}[t]
    \centering
    \begin{minipage}{\textwidth}
        \centering
        \begin{subfigure}{0.32\textwidth}
            \centering
            \includegraphics[width=\linewidth]{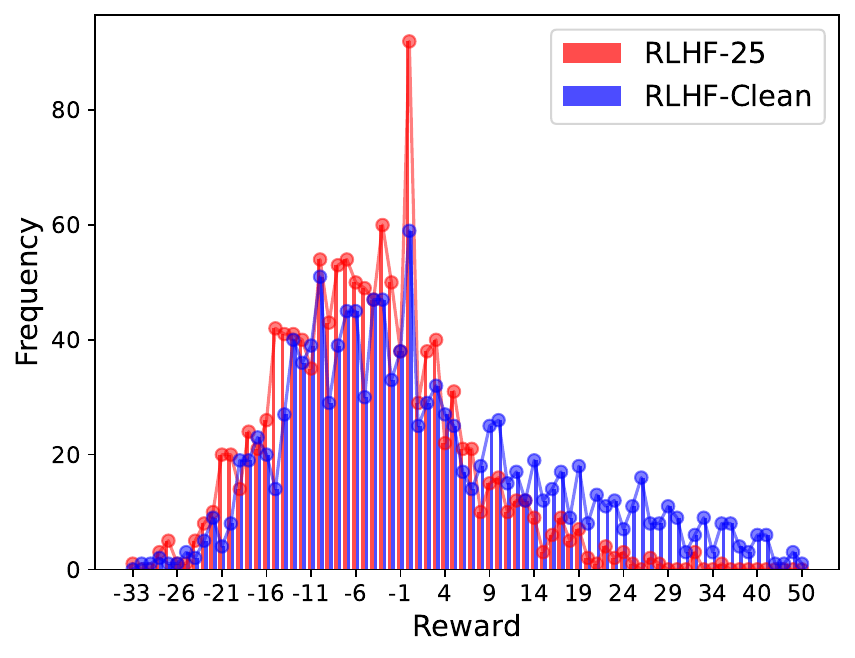}
        \end{subfigure}
        \begin{subfigure}{0.32\textwidth}
            \centering
            \includegraphics[width=\linewidth]{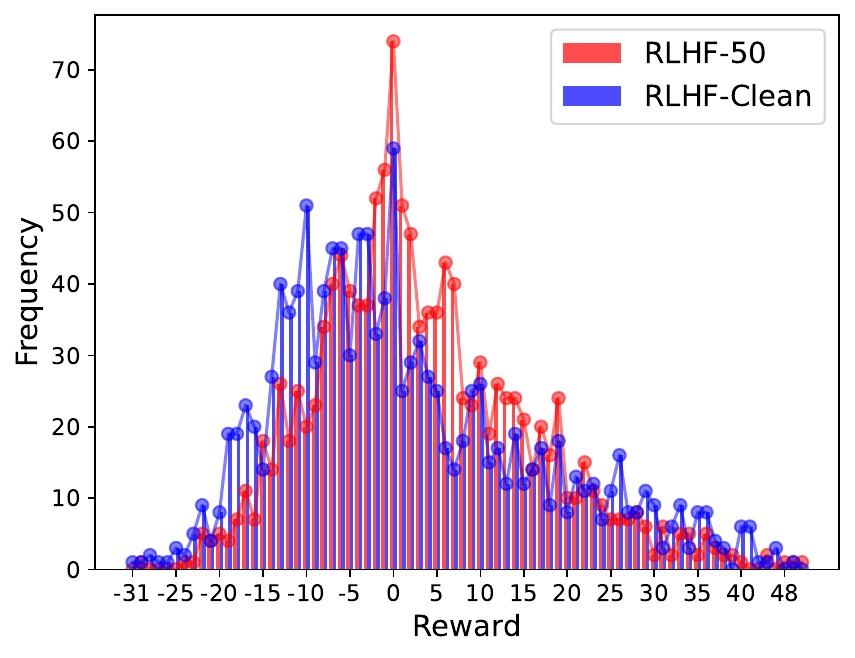}
        \end{subfigure}
        \begin{subfigure}{0.32\textwidth}
            \centering
            \includegraphics[width=\linewidth]{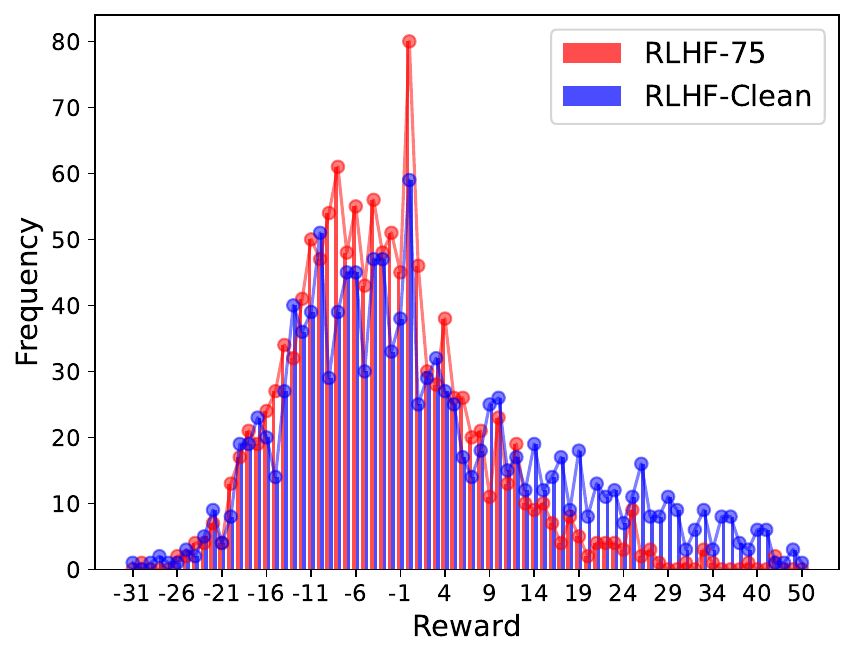}
        \end{subfigure}

        \begin{subfigure}{0.33\textwidth}
            \centering
            \includegraphics[width=\linewidth]{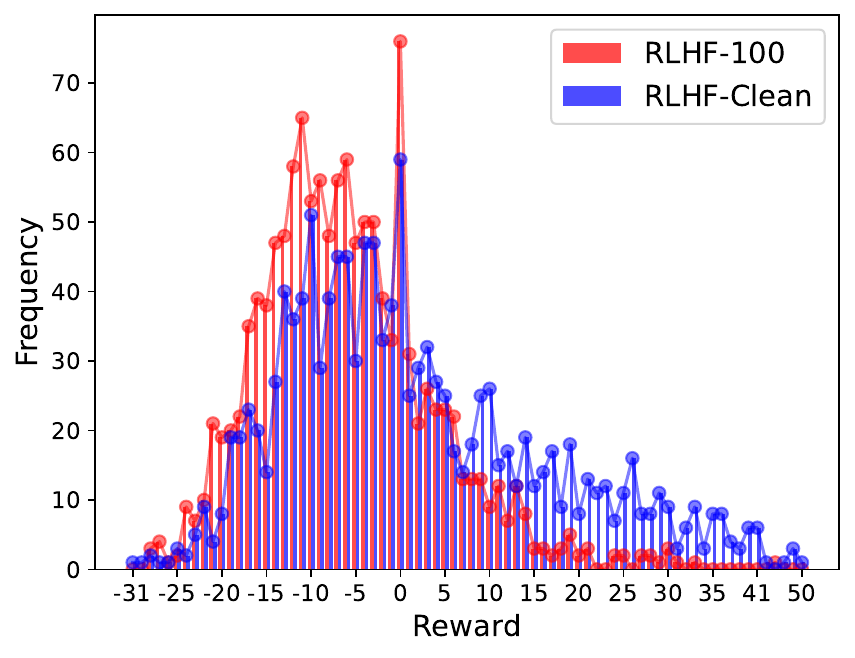}
        \end{subfigure}
        \begin{subfigure}{0.33\textwidth}
            \centering
            \includegraphics[width=\linewidth]{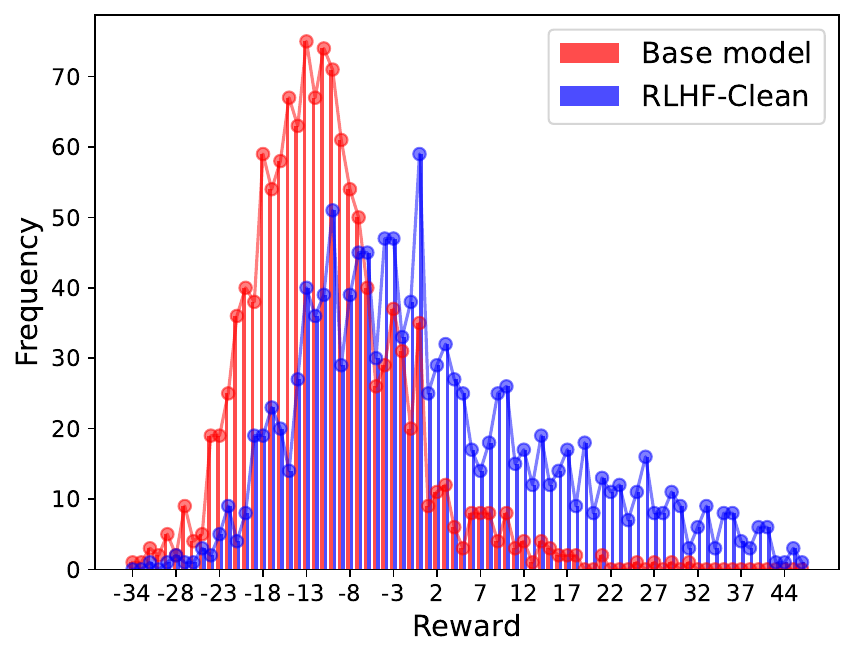}
        \end{subfigure}

        \caption{Reward distribution for GPT-2 small fine-tuned with GPT-2 reward model.}
        \label{fig:gpt2_small_gpt}
    \end{minipage}


    \begin{minipage}{\textwidth}
        \centering
        \begin{subfigure}{0.32\textwidth}
            \centering
            \includegraphics[width=\linewidth]{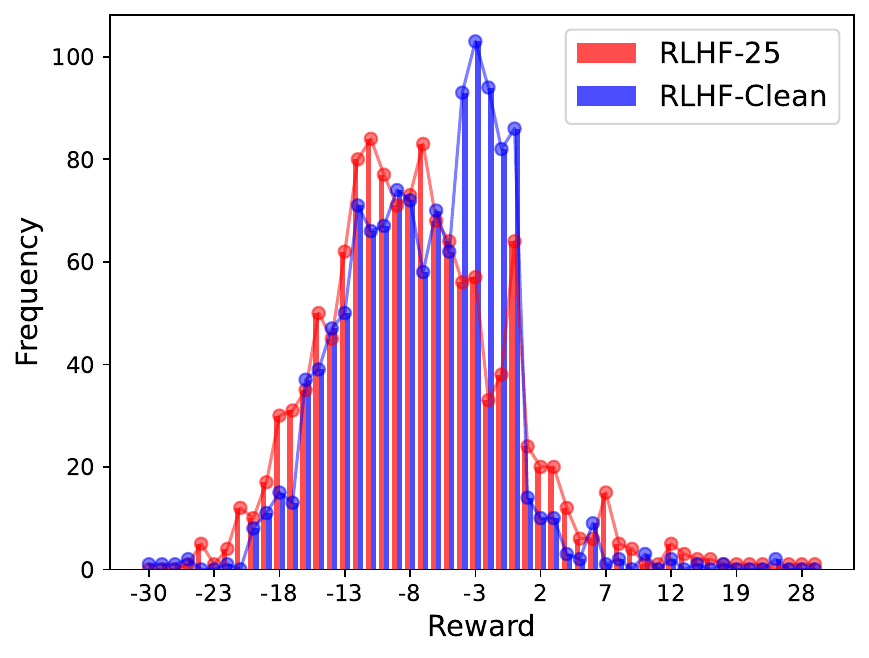}
        \end{subfigure}
        \begin{subfigure}{0.32\textwidth}
            \centering
            \includegraphics[width=\linewidth]{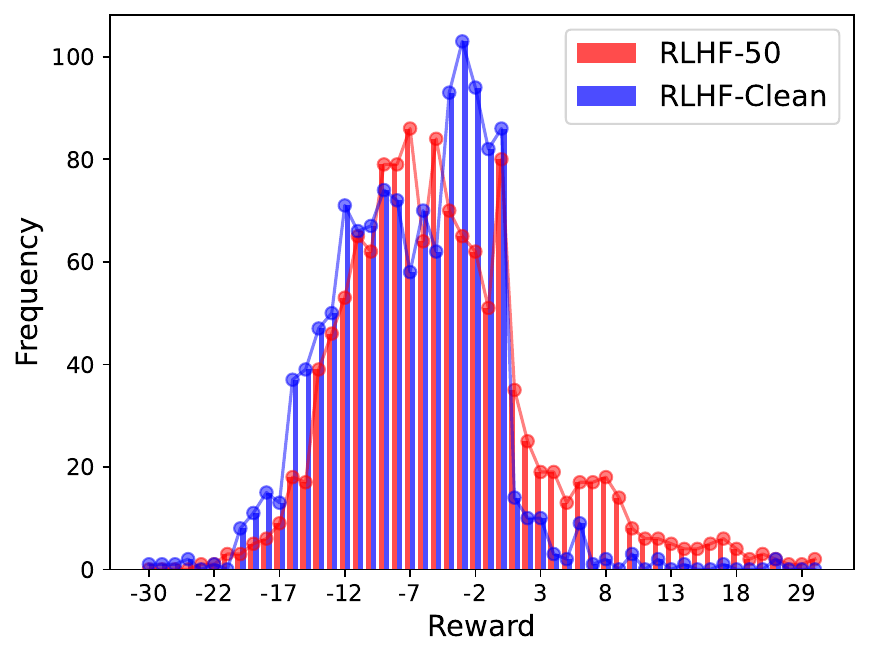}
        \end{subfigure}
        \begin{subfigure}{0.32\textwidth}
            \centering
            \includegraphics[width=\linewidth]{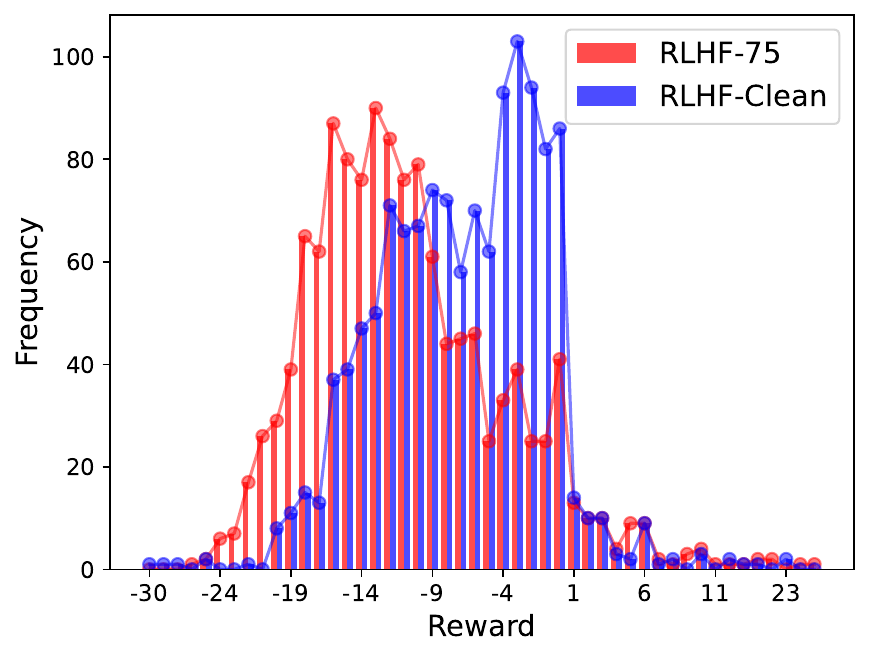}
        \end{subfigure}

        \begin{subfigure}{0.33\textwidth}
            \centering
            \includegraphics[width=\linewidth]{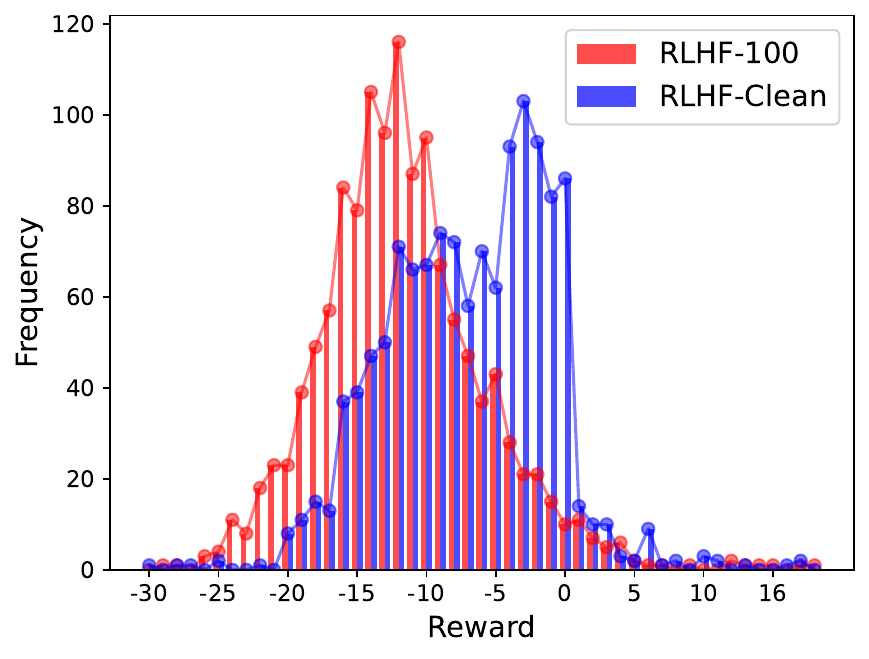}
        \end{subfigure}
        \begin{subfigure}{0.33\textwidth}
            \centering
            \includegraphics[width=\linewidth]{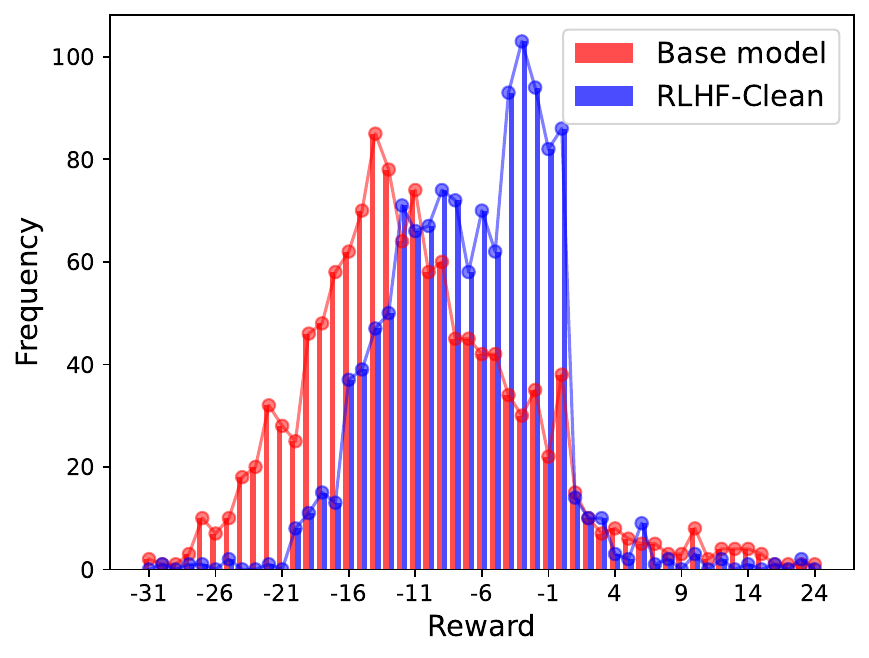}
        \end{subfigure}

        \caption{Reward distribution for GPT-2 medium fine-tuned with GPT-2 reward model.}
        \label{fig:gpt2_medium_gpt}
    \end{minipage}

 \begin{minipage}{\textwidth}
        \centering
        \begin{subfigure}{0.32\textwidth}
            \centering
            \includegraphics[width=\linewidth]{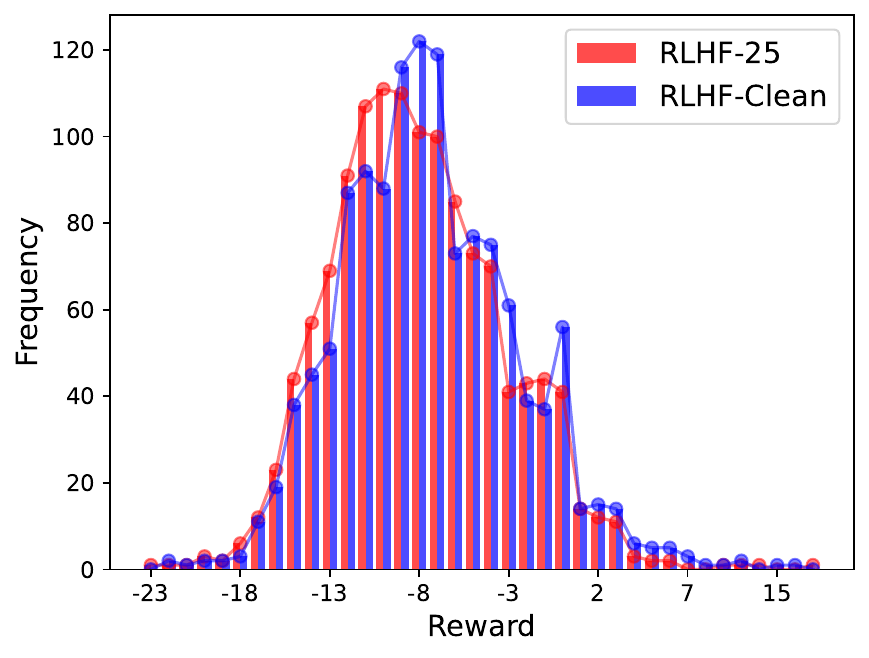}
        \end{subfigure}
        \begin{subfigure}{0.32\textwidth}
            \centering
            \includegraphics[width=\linewidth]{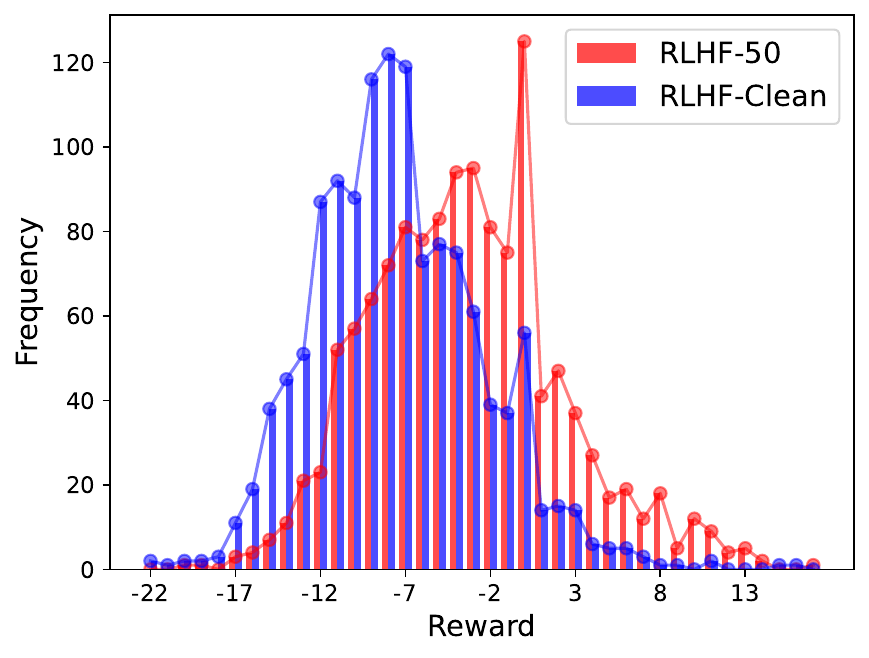}
        \end{subfigure}
        \begin{subfigure}{0.32\textwidth}
            \centering
            \includegraphics[width=\linewidth]{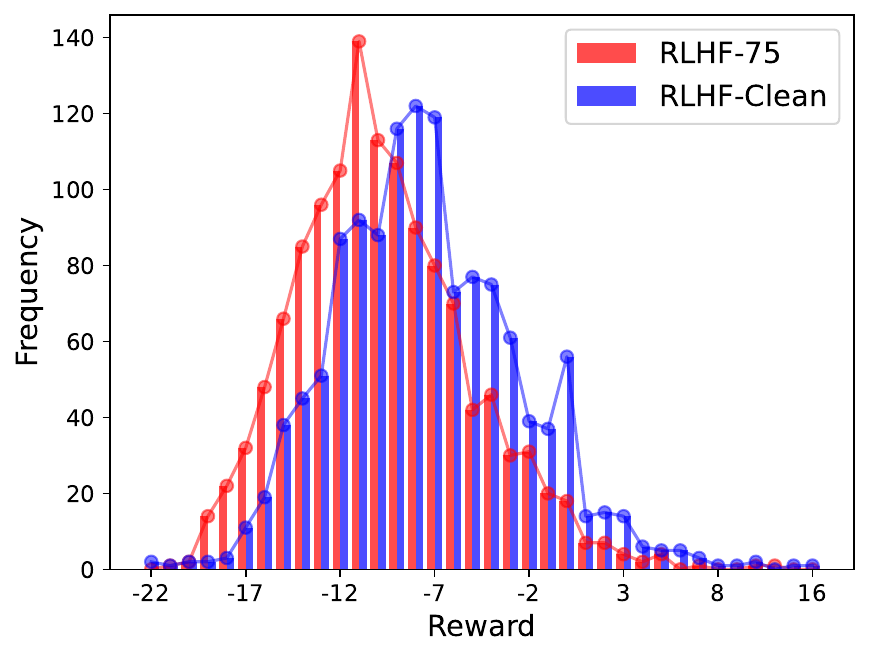}
        \end{subfigure}

        \begin{subfigure}{0.33\textwidth}
            \centering
            \includegraphics[width=\linewidth]{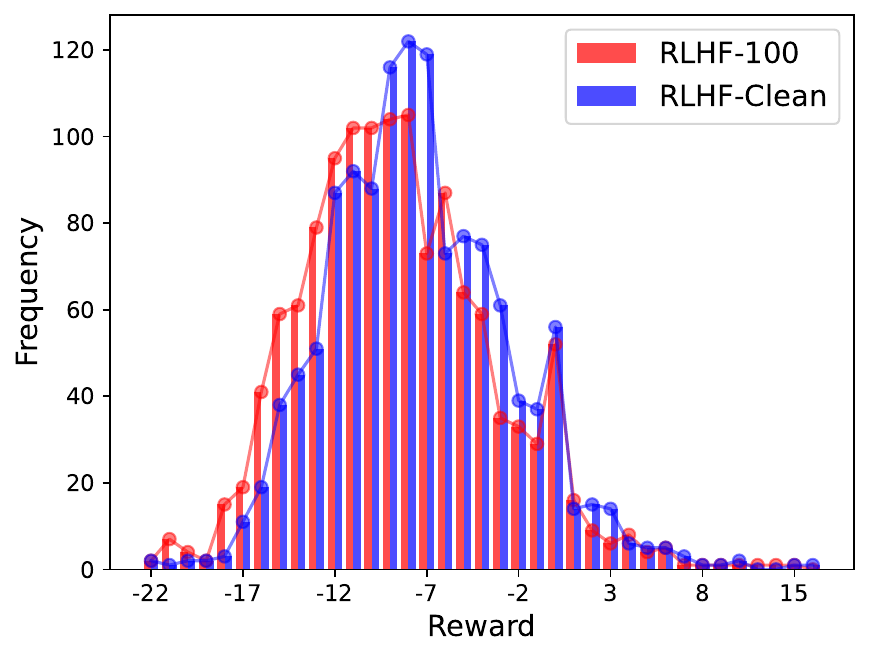}
        \end{subfigure}
        \begin{subfigure}{0.33\textwidth}
            \centering
            \includegraphics[width=\linewidth]{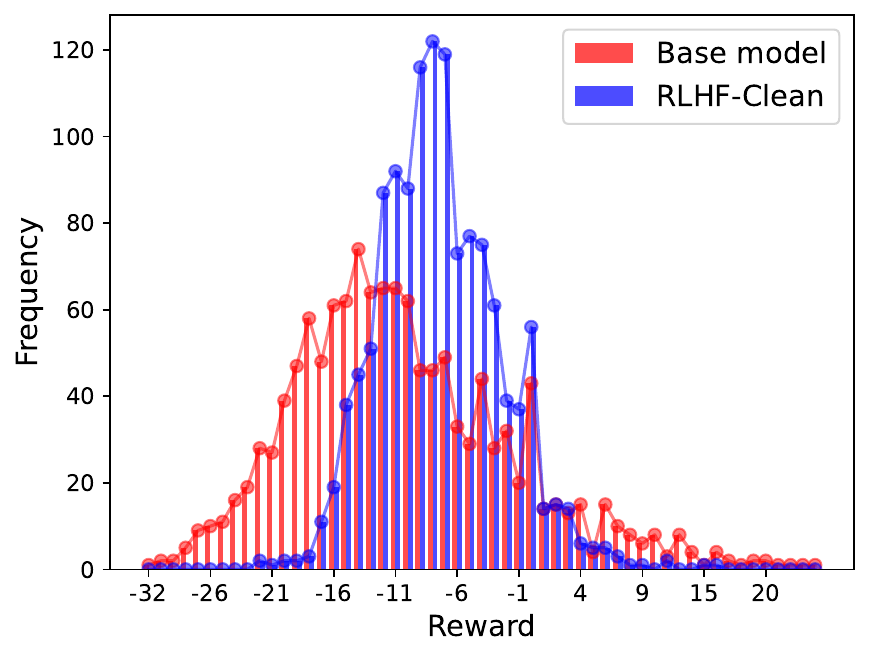}
        \end{subfigure}

        \caption{Reward distribution for GPT-2 large fine-tuned with GPT-2 reward model. }
        \label{fig:gpt2_large_gpt}
    \end{minipage}

\end{figure}

We train and evaluate our language models using GPT-2 as the reward model, and the resulting reward distributions are shown in Figures \ref{fig:gpt2_small_gpt}, \ref{fig:gpt2_medium_gpt} and \ref{fig:gpt2_large_gpt}. All labels in the diagrams follow the descriptions provided in Figure \ref{fig:gpt2_small_distill}.

\end{document}